%% file: collas2024_conference.tex
\documentclass{article} 
\usepackage{collas2024_conference,times}
\usepackage{easyReview}

\usepackage[fleqn]{amsmath}
\usepackage{algorithm}
\usepackage{algpseudocode}

\input{math_commands.tex}

\usepackage{hyperref}
\hypersetup{
    colorlinks=true,
    linkcolor=red,
    filecolor=magenta,
    urlcolor=blue,
    citecolor=purple,
    pdftitle={Overleaf Example},
    pdfpagemode=FullScreen,
    }

\title{Simplified priors for Object-Centric Learning}

\author{
\vspace{0.5cm}
 \hspace{2.5cm}
  Vihang Patil$^{*1}$
  Andreas Radler$^{* \ 1}$
  \textbf{Daniel Klotz}$^{\ 2}$
  \textbf{Sepp Hochreiter}$^{\ 1,3}$\\
\vspace{0.1cm}
\hspace{2.6cm}
{$^*$}{Equal contribution; order randomly assigned}\\ 
\vspace{0.1cm}
\hspace{2.7cm}{$^1$}{LIT AI Lab, Institute for Machine Learning, JKU Linz, Austria}\\
\vspace{0.1cm}
\hspace{2.7cm}{$^2$}{Helmholtz Centre for Environmental Research (UFZ), Leipzig, Germany}\\
\vspace{0.1cm}
\hspace{2.7cm}{$^3$}{NXAI GmbH, Linz, Austria}\\
\hspace{2.7cm}\texttt{\{patil, radler, klotz, hochreit\}@ml.jku.at}
}

\collasfinalcopy

\begin{document}

\maketitle

\begin{abstract}
Humans excel at abstracting data and constructing \emph{reusable} concepts, a capability lacking in current continual learning systems. 
The field of object-centric learning addresses this by developing abstract representations, or slots, from data without human supervision.
Different methods have been proposed to tackle this task for images, whereas most are overly complex, non-differentiable, or poorly scalable.
In this paper, we introduce a conceptually simple, fully-differentiable, non-iterative, and scalable method called \textbf{SAMP} (\textbf{S}implified Slot \textbf{A}ttention with \textbf{M}ax Pool \textbf{P}riors). 
It is implementable using only
Convolution and MaxPool layers and an Attention layer.
Our method encodes the input image with a Convolutional Neural Network and then uses a branch of alternating Convolution and MaxPool layers to 
create specialized sub-networks and 
extract primitive slots. 
These primitive slots are then used as queries for a Simplified Slot Attention over the encoded image.
Despite its simplicity, our method is competitive or outperforms previous methods on standard benchmarks.
\end{abstract}

\section{Introduction}


In a dynamic environment, it is essential for artificial intelligence systems to be able to effectively adapt and generalize to similar unseen input.
The field of continual learning aims to design adaptable systems, while mitigating forgetting of important previously acquired capabilities (i.e. catastrophic forgetting).
Many current methods approach continual learning via memory stability, or inter-task generalizability mechanisms \citep{wang2024LifeLongLearningReview}, while being agnostic to the underlying representation. 
Recently object-centric learning has gained momentum as a promising direction to build learning systems inspird by human cognition \citep{SlotAttention,IODINE,Monet,greff-Binding_problem}. 
The core idea is, that an object-centric representation of the raw-input may lead to better generalization and adaptation capabilities.
These symbolic-like representations for objects, can then
be applied, reused, and combined for reasoning or decision-making \citep{Spelke2007-SPECK-3-dev_of_5_year_old}.
However, despite the rise of Deep Learning \citep{Krizhevsky:12, Schmidhuber:15} and access to large scale compute and data \citep{Deng:09}
that led to human level performance in various domains \citep{He:16, Mnih:15, Silver:16}, neural networks do not posses this abstraction ability. 

Various recent works have tried to
address the problem of abstraction from raw input by learning object-centric representations --- often known as slots \citep{SlotAttention, wu2023slotformer, ImplicitSlotAttention, Genesis, Monet} --- from high-dimensional input, such as images, videos, 3D point clouds or even 3D scenes (images of different angles onto the same objects).
However, most of the currently proposed methods are complex and partially suffer from training instabilities \citep{ImplicitSlotAttention}.
Many of these methods are derived from the original Slot Attention idea \citep{SlotAttention}, which applies a modification of Cross-Attention \citep{Transformer_Vaswani} in an iterative refinement procedure. 
We argue that such an iterative refinement procedure is not desired and propose a solution that can serve as a simple baseline.

Prior to explicit slot extraction methods, \emph{competition mechanisms} \citep{Srivastava:13_compete_to_compute} like MaxPooling or Spatial Pyramid MaxPooling \citep{spatial_pyramid_pooling} showed a strong performance \citep{ResNet, 
mask_rcnn, InceptionNet} to obtain highly relevant abstract features.
\citet{srivastava2014understanding_locally_comp_nets} empirically showed that competition among subnetworks leads to specialization, whereas for different input, different subnetworks are activate.
We argue that this ability to specialize and abstract are useful properties for learning object-centric representations.

In this paper, we present a novel method that we call Simplified Slot Attention with Max Pool Priors (SAMP). SAMP is a simple, scalable, non-iterative and fully-differentiable approach to extract slots from images with simple Convolutions \citep[CNN;][]{LeCun:04} and MaxPool blocks \citep{Krizhevsky:12}  that are
combined with a Simplified Slot attention (SSA) layer.
The effectiveness of SAMP demonstrates that the iterative nature of Slot Attention based methods is not necessary.

\hfill

Our main contributions are:
\begin{itemize}
    \item We propose SAMP (Simplified Slot Attention with Max Pool Priors) a method for learning object-centric representations.     
    \item SAMP is a simple, but scalable baseline for OCL, since it is non-iterative and consists of
    vanilla building blocks like CNN, MaxPool layers and a Simplified Slot-Attention. 
    \item We evaluate SAMP on standard Object-Centric benchmarks, where it is competitive or outperforms various other Object-Centric methods. 
\end{itemize}

\begin{figure*}[t]
    \centering
    \includegraphics[width=\linewidth]{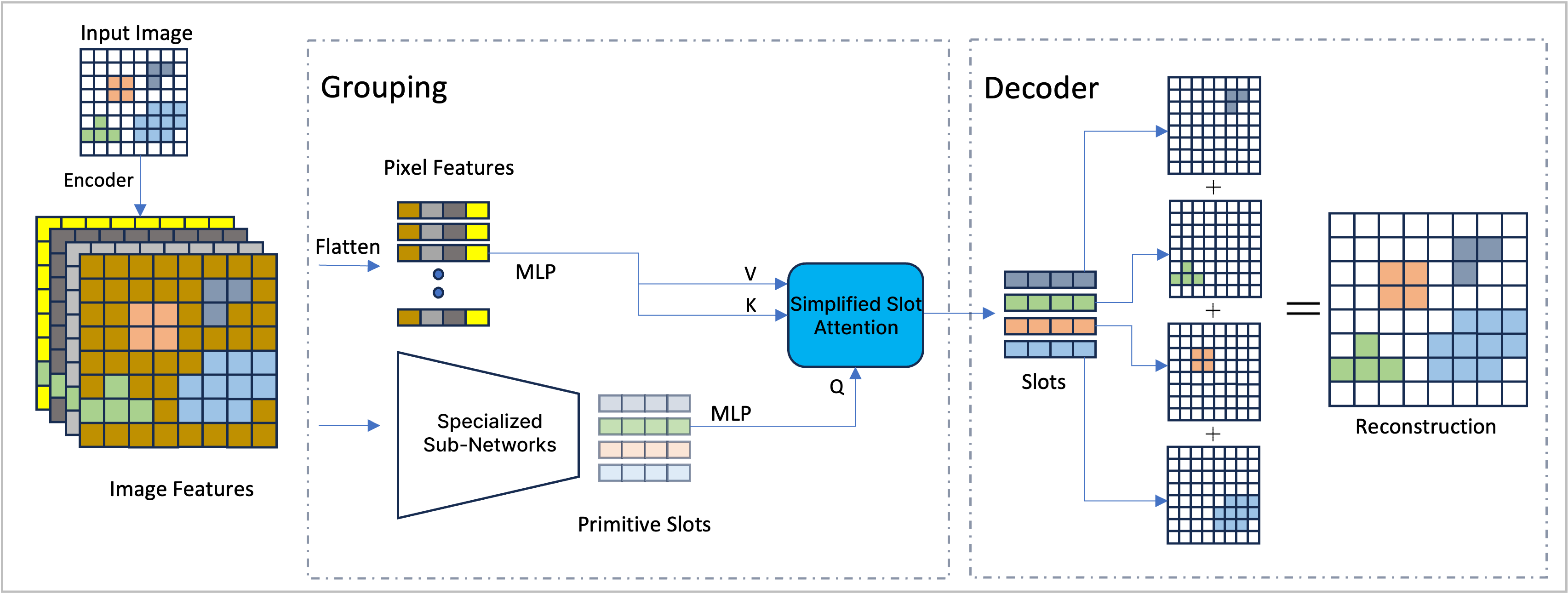}
    \caption{\textit{Grouping:} We learn Primitive Slots from image features using Specialized Sub-Networks. We obtain pixel features by flattening all the image features from the encoder. 
    We pass pixel features and Primitive Slots to a Simplified Slot Attention (SSA) layer, where \textit{Keys} (K) and \textit{Values} (V) are the pixel features and \textit{Queries} are the Primitive Slots. SSA layer outputs the slots. The decoder is applied on every slot separately to reconstruct the input and a mask. 
    A softmax is applied to the masks along the pixel dimension (for simplicity the masks are not shown in the figure). 
    The final reconstruction is obtained by performing a weighted sum of all the individual reconstructions across the pixels with the weights coming from the masks.}
    \label{fig:grouping}
\end{figure*}

\begin{figure}[t]
    \centering
    \includegraphics[width=0.8\linewidth]{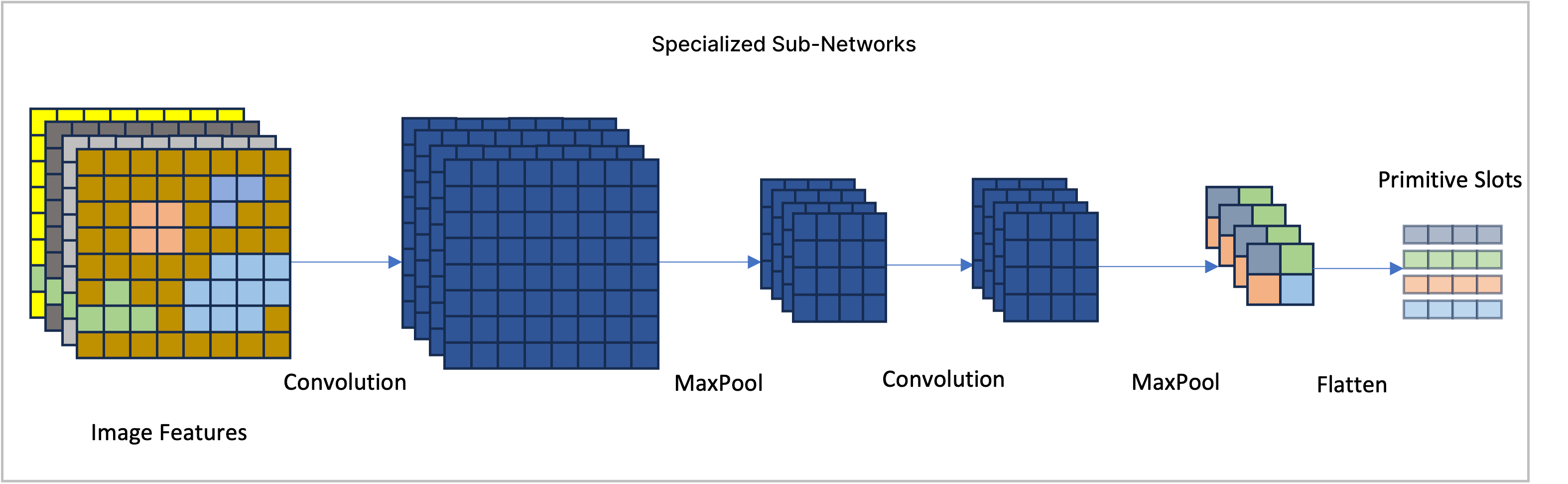}
    \caption{\textit{Specialized Sub-Networks:} We use alternating Convolution and MaxPool layers. After these layers, we flatten features to obtain Primitive Slots. The architecture along with the slot-wise reconstruction in the decoder, induces specialization in sub-networks. The sub-networks are forced to explain different parts of the input. Therefore, the resultant Primitive Slots are good queries for the SSA layer.}
    \label{fig:comp_subnets}
\end{figure}

\begin{minipage}[t]{0.46\textwidth}
\begin{algorithm}[H]
    \centering
    \caption{Simplified Slot Attention (SSA) Layer: SSA removes the iteration from Slot Attention, along with GRU and the MLP layers. It also has shared projection weights for keys and values. Further, we do not normalize across keys after the softmax over slots.}\label{algorithm1}
    \footnotesize
    \begin{algorithmic}[1]
        \State \text{\textbf{Input}: inputs $\in \mathbb{R}^{P \times D}$, slot\_priors $\in \R^{n \times D}$}
        \State \textbf{Layer Params}: k (shared projection for keys and values), q: projection for slot priors; LayerNorm(x2)
        \State $inputs = LayerNorm(inputs)$
        \State $slots = LayerNorm(slot\_priors)$
        \State $M = \frac{1}{\sqrt{n}}k(inputs)\cdot q(slots)^{T}$ 
        \State $W_{i,j} = \frac{e^{M_{i,j}}}{\sum_{l=1}^{n}e^{M_{l,j}}}$ 
        \State $slots = W^{T}\cdot k(inputs)$
        \State \textbf{return} slots
    \end{algorithmic}
\end{algorithm}
\end{minipage}
\hfill
\begin{minipage}[t]{0.46\textwidth}
\begin{algorithm}[H]
    \centering
    \caption{Slot Attention Module as given in \cite{SlotAttention}}\label{algorithm2}
    \footnotesize
    \begin{algorithmic}[1]
        \State \text{\textbf{Input}: inputs $\in \mathbb{R}^{P \times D}$, slots $\sim \mathcal{N}(\mu, diag(\sigma)) \in \R^{n \times D}$}
        \State \textbf{Layer Params}: k, q, v: linear projections for attention; GRU; MLP; LayerNorm(x3)
        \State $inputs = LayerNorm(inputs)$
        \For{$t = 0....T$}
        \State $slots\_prev = slots$
        \State $slots = LayerNorm(slots)$
        \State $M = \frac{1}{\sqrt{D}}k(inputs)\cdot q(slots)^{T}$
        \State $attn_{i,j} = \frac{e^{M_{i,j}}}{\sum_{l=1}^{n}e^{M_{i,l}}}$
        \State $W_{i,j} = \frac{attn_{i,j}}{\sum_{l=1}^{P}attn_{l, j}}$
        \State $updates = W^{T}\cdot v(inputs)$
        \State slots = GRU(state=slots\_prev, inputs=updates)
        \State slots += MLP(LayerNorm(slots))
        \EndFor
        \State \textbf{return} slots
    \end{algorithmic}
\end{algorithm}
\end{minipage}

\section{Related Work}

Object-centric learning is related to abstraction, which is a well studied topic in different areas of Machine Learning \citep{Givan:03, Ravindran:03b, Sutton:99, Li:06, Vezhnevets:17, Kulkarni:16, patil:22, Patil:23}. 
Object-centric learning methods try to extract representations of objects given a input, which could be images or some other modality.
These extracted representations are known as slots.

Many such methods consist of three modules: an image encoder, a grouping module and a decoder.
The image encoders are typically realized with multiple CNN layers \citep{SlotAttention, GenesisV2}. For the decoder, the spatial broadcast decoder \citep{watters2019SBD} --- or slight deviations thereof --- is the de-facto standard, since it provides a good inductive bias for learning disentangled representations \citep{engelcke2020reconstruction}.
The main difference between common slot extraction methods is the \emph{grouping module}. 
Most grouping modules can be divided into 
(a) graph-based, (b) generative and (c) iterative refinement based approaches.

\paragraph{Graph-based approaches}
\citet{pervez2022graphcut_slots} use a graph-cutting approach to perform clustering of pixel-features via a quadratic program.
The disadvantage of this approach is that solving the underlying quadratic program is computationally intense and poorly parallelizable. 

\paragraph{Generative approaches} aim to learn  parameterized distributions of latent variables by minimizing a proxy of the log-likelihood.
Works such as IODINE \citep{IODINE}, MONet \citep{Monet}, SPACE \citep{SPACE_method} or Genesis \citep{Genesis} fall into this category. While these methods are theoretically grounded, they lack empirical performance.
Genesis-V2 \citet{GenesisV2} is an extension of Genesis and one of the first fully-differentiable methods that is able to choose the number of slots \emph{by design} during training and inference (i.e., not by doing a second forward pass with a different number of slots). Although this selection option brings some interesting properties, the method has not found wide adoption. We argue that the reason for this is that it is complex in terms of math and code.

\paragraph{Iterative refinement approaches}
The Slot Attention module \citep{SlotAttention} --- often simply called Slot Attention ---is a successful and widely adapted technique.
Thus, there exists a sizable body of work that is dedicated to improve it: Implicit Slot Attention \citep[ISA][]{ImplicitSlotAttention} stabilizes the iterative refinement process of Slot Attention by learning the fixed-point of the iterative refinement procedure with a first-order Neumann series approximation. 
While it stabilizes the training, it tends to need even more iterations than Slot Attention which increases the overall time complexity.
\citet{jia2023_ISA_w_query_optimization} extend Implicit Slot Attention by learning distributions for initializations of the iterative refinement procedure.
An other improvement suggestion to the original Slot Attention module came from \citep{kim2023shepherding} who introduced a locality-bias as it is common for CNNs in the attention part of Slot Attention.
Similarly, \citet{gao2023_SA_with_cluster_init} improved the slot initialization with explicit clustering methods, such as Mean-Shift \citep{Mean_Shift_Clustering} or K-means \citep{K_means_clustering}.
SLATE \citep{SLATE_singh2022} uses Slot Attention as the grouping module, but replaces CNN encoder with a discrete Variational Autoencoder \citep{dVAE} encoder and the CNN spatial broadcast decoder \citep{watters2019SBD} with an autoregressive transformer decoder \citep{Transformer_Vaswani}.


\paragraph{Competition in Neural Networks}
Competition in artificial neural networks is inspired by observations of the human brain.
Competitive dynamics among neurons and neural circuits have played a crucial role in understanding brain processes in biological models \citet{Eccles:67, Ermentrout:92}. 
The significance stems from early findings that revealed a recurring "on-center, off-surround" neural architecture in various brain regions \citet{Ellias:75}. 
The "on-center, off-surround" architecture involves neurons providing excitatory feedback locally and inhibitory signals broadly. Therefore it creates competition between sub-networks \citep{Andersen:69, Ellias:75}. 
The biological model in \citet{Lee:99} is also of interest,
as it proposes that attention activates a winner-take-all competition
over the visual features in the human brain. 
Inspired from the brain, such competitive networks have made there way to machine
learning \citep{Maass:99, Maass:00}. 
Hamming Networks \citep{Lippmann:87} perform iterative lateral inhibition
between competitors.
Local-Winner-Take-All networks \citep{Srivastava:14} setup blocks of neurons, 
and pass the winning activation of each block to the next layer. 
Adding sources of competition on
networks trained on faces, showed impressive abstraction capabilities, where feature maps of deeper layers (i.e., closer to the output) become more abstract and depict parts of a face \citep{wang2014_face_more_abstract_features}.


\section{Method}
\label{sec:method}
We propose \textit{Simplified Slot Attention with Max Pool Priors}, a method to learn Object-Centric representations from images. 
Most architectures consist of three main modules: an encoder, a grouping and a decoder module.
The encoder computes useful features from the raw image, the grouping module extracts object-representations (slots) and the decoder reconstructs the individual slots to partial images, which are then summed up to obtain the input image in an auto-encoder fashion.
The main novelty of SAMP lies in its simplified grouping mechanism (Figure \ref{fig:grouping} and \ref{fig:comp_subnets}). 
SAMP groups image features using specialized sub-networks
and a variant of the slot attention layer. 
Competition and specialization is induced in SAMP due to the following components: 
1) Winner Take All MaxPool layers 2) Simplified Slot Attention layer
3) Spatial Broadcast Decoder. 

In the following sub-section (\ref{subsec:architecture}), we first go through the architecture of SAMP.
Afterwards, we describe the sources of competition and specialization in SAMP (sub-section \ref{subsec:comp_in_SAMP}). 




\subsection{Architecture}
\label{subsec:architecture}
The SAMP architecture consists of an encoder, grouping module and a decoder. 

\paragraph{Encoder} 
The encoder is a simple image encoder with CNN layers \citep{LeCun:04}. 
We have a fixed kernel size, padding and stride across these layers. 
The layers are setup in such a way that the spatial dimension of the input is preserved.
For example, if the input image is of size $H \times W$ then the encoder output is $H \times W \times c$,
where, $H$ and $W$ are the height and width of the image. 
$c$ is the number of filters used in the last layer. 
This is done to obtain features at pixel level, which can then be later grouped together to obtain slots.  
Similar to \citet{SlotAttention}, we also augment the pixel features with a positional embedding. 
For more details, see Appendix \ref{app:hyperparameters}.

\paragraph{Grouping}
The output of the encoder is fed to SAMP's grouping module, which outputs slots. 
The grouping module consists of specialized sub-networks 
and a SSA Layer. 

The specialized sub-networks ingest the encoder output and return primitive-slots. 
A specialized sub-networks consist of MaxPool and Convolution layers. 
Similar to the encoder, we build the CNN layers in such a way that they preserve the spatial dimension. 
But, the MaxPool layers reduce it (Figure \ref{fig:comp_subnets}). 
After applying CNN and MaxPool layers we obtain an output of size $n_h \times n_w \times c$. Here,
$n_h, n_w$ are height and width of the output and $c$ is the number of filters of the last CNN layer.
We flatten the output to obtain, $n \times c$, where $ n = n_h \times n_w$ are the number of 
primitive slots. 
Conceptually, $n$ is therefore a hyperparameter of the network. 
We pass these primitive-slots through an MLP layer and then use them as queries in the SSA Layer. 


The SSA Layer is a variant of the standard slot attention layer \citep{SlotAttention}. 
We remove the multiple iterations done in slot attention along with the Gated recurrent unit (GRU) layer \citep{ChungGCB14_GRU} and Multilayer perceptron (MLP) layer.
We also share projection weights between the keys and values, thus, keys and values are same in our architecture. 
Unlike, slot attention, after the softmax over the queries, we do not normalize attention coefficients across the keys. 
The softmax over the queries
creates competition in the slots to explain different parts of the input. 
In our case, as the keys and values share the same projection weights it can also be thought of as an associative memory.
In that case it is 
a storage of patterns \citep{Ramsauer:20, Widrich:21, Paischer:22},
from which we retrieve patterns (values) similar to primitive slots (queries). 

SAMP first flattens the encoder output and passes it through MLPs so they can be used as keys ($\mathbf{K} \in \mathbb{R}^{P \times D}$) and values ($\mathbf{V} \in \mathbb{R}^{P \times D}$)
in the SSA Layer layer, whereas $P$ is the number of pixels and $D$ is the dimensionality of the slots. 
The primitive-slots are used as queries ($\mathbf{Q} \in \mathbb{R}^{n \times D}$). Once we get the attention coefficients, we can use them to get the slots ($\mathbf{S} \in \mathbb{R}^{n \times D}$) as follows by using 
\begin{equation}
    \qquad \qquad \qquad \qquad \qquad \qquad \qquad \qquad \mathbf{S} = \mathbf{W}^{T} \mathbf{V} \qquad \qquad \qquad \qquad \text{  where, } \mathbf{W} = \mathrm{softmax}(\frac{\mathbf{K} \mathbf{Q}^{T}}{\tau})
\end{equation}

The temperature $\tau$ is set to $\sqrt{n}$, whereas $n$ is the number of slots. This is a lower temperature then commonly used in Cross-Attention \citep{Transformer_Vaswani}. Empirically we found that the resulting lower-entropy distribution of attention values leads to better results.


\paragraph{Decoder}
The decoding treats each slot separately with a Spatial Broadcast Decoder \citep{IODINE}. 
Every slot reconstructs its own image
and a mask. 
The masks are normalized and a softmax along the pixel dimension of all masks 
mixes all the images together. 
We train the model end to end with a mean squared error reconstruction loss \citep[similar to][]{IODINE, SlotAttention}. 

Our method is able to extract slots, without any iterative refinement. 
In the next subsection, we explain how 
SAMP creates competition amongst the slots to explain different parts of the input.

\subsection{Competition in SAMP}
\label{subsec:comp_in_SAMP}

SAMP creates competition and specialization through different sources: The MaxPool Layers, the SSA Layer layer and a slot wise reconstruction decoder.  

\paragraph{Specialization through MaxPool layers}
MaxPool layers are interleaved between
Convolution layers (Figure \ref{fig:comp_subnets}). 
MaxPool layers work by transmitting activations of winning units to the next layer.
As there is only one winner for a local group of neurons, 
the layer activations are sub-sampled. 
This results in sub-networks competing to have higher activations. 
During back-propagation, units that win 
and the subnetworks that are responsible for this will get updated. 
As a result, a winning sub-network is reinforced to win more if it predicts correctly. 
If a sub-network wins, and does not predict correctly, it is not reinforced. 
One can also look at this as a gradient-based search over finding sub-networks, which explain the input correctly. 
An important side effect of these specialized sub-networks
is that the resulting primitive-slots explain different parts of the input. 
A neuron has a higher chance of winning if it explains a different part of 
the input, 
rather than explaining the same feature as another neuron. 

\paragraph{Competition through SSA Layer}
Another source of competition for SAMP is the SSA Layer layer. 
The layer takes a softmax across queries,
this forces the queries to compete for keys (pixel features). 
We see improvement in results by using this layer, 
over not using it (See Table \ref{table:ablations}). 

\paragraph{Competition through Spatial Broadcast Decoder}
Finally, we use a slot wise reconstruction decoder, 
i.e., every slot is separately fed to the decoder for reconstruction (Figure \ref{fig:grouping}).
The final reconstruction is obtained by mixing all reconstructions with
a softmax over the respective masks. 
This creates competition over pixels in the final reconstructed image. 
Due to the mixing of the reconstructions, 
the slots get reinforced for explaining different parts of the input. 
Thus, providing a push to the competitive nature of the slots. 
This has further been discussed in \citet{engelcke2020reconstruction}.

In summary, MaxPool layers, SSA Layer layer and the decoder lead to competition and specialization amongst the slots and as a result they try to explain different parts of the input.

\subsection{Advantages}

The main advantage of our method over Slot Attention is, that it is non-iterative and therefore provides a better time and space complexity (see table \ref{complexities-table}).
%
\begin{table}[h]
\caption{Runtime and Memory Complexities of the grouping module of Slot Attention vs. SAMP.
    $S$ denotes the number of slots, $D$ is the dimensionality of the slots and the number of channels (these are assumed to be equal for simplicity), $H$ and $W$ are the height and width of the image.
    While SAMP uses another few layers for the competition, Slot Attention performs several iterations of the iterative refinement procedure.
    }
\label{complexities-table}
\begin{center}
\begin{tabular}{lcc}
\multicolumn{1}{c}{\bf }  &\multicolumn{1}{c}{\bf Slot Attention}  &\multicolumn{1}{c}{\bf SAMP (ours)} \\
\hline
Time (Training) & $\mathcal{O}(T \cdot S \cdot H \cdot W \cdot D)$ & $\mathcal{O}((L_{\text{comp}} + S) \cdot H \cdot W \cdot D)$ \\
Space (Training) & $\mathcal{O}(T \cdot S \cdot H \cdot W \cdot D)$ & $\mathcal{O}((L_{\text{comp}} + S) \cdot H \cdot W \cdot D)$ \\
Time (Test) & $\mathcal{O}(T \cdot S \cdot H \cdot W \cdot D)$ & $\mathcal{O}((L_{\text{comp}} + S) \cdot H \cdot W \cdot D)$ \\
Space (Test) & $\mathcal{O}(S \cdot H \cdot W \cdot D)$ & $\mathcal{O}((L_{\text{comp}} + S) \cdot H \cdot W \cdot D)$ \\
\end{tabular}
\end{center}
\end{table}
\paragraph{Scalability}

The Slot Attention module is an \textbf{iterative refinement method} that needs several iterations to converge to a fixed point. 
As the complexity of the task increases, so does the number of required iterations.
\citet{ImplicitSlotAttention},
while addressing some of the training instabilities of Slot Attention, require even more iterations. In the case of training on CLEVR, \citet{ImplicitSlotAttention} may utilize up to 11 iterations. These iterative methods, however, pose a significant challenge when it comes to scaling them for large-scale datasets.

SAMP
distinguishes itself by not relying on recurrent neural networks (RNNs) and being a non-iterative method. This characteristic makes SAMP particularly well-suited for scaling up to handle large datasets and computational demands efficiently. SAMP's non-iterative nature simplifies its application to extensive data processing and computation tasks, providing a distinct advantage over iterative methods like \citet{SlotAttention} and \citet{ImplicitSlotAttention}.




\section{Experiments}

\begin{figure}[t]
    \centering
    \includegraphics[width=\linewidth]{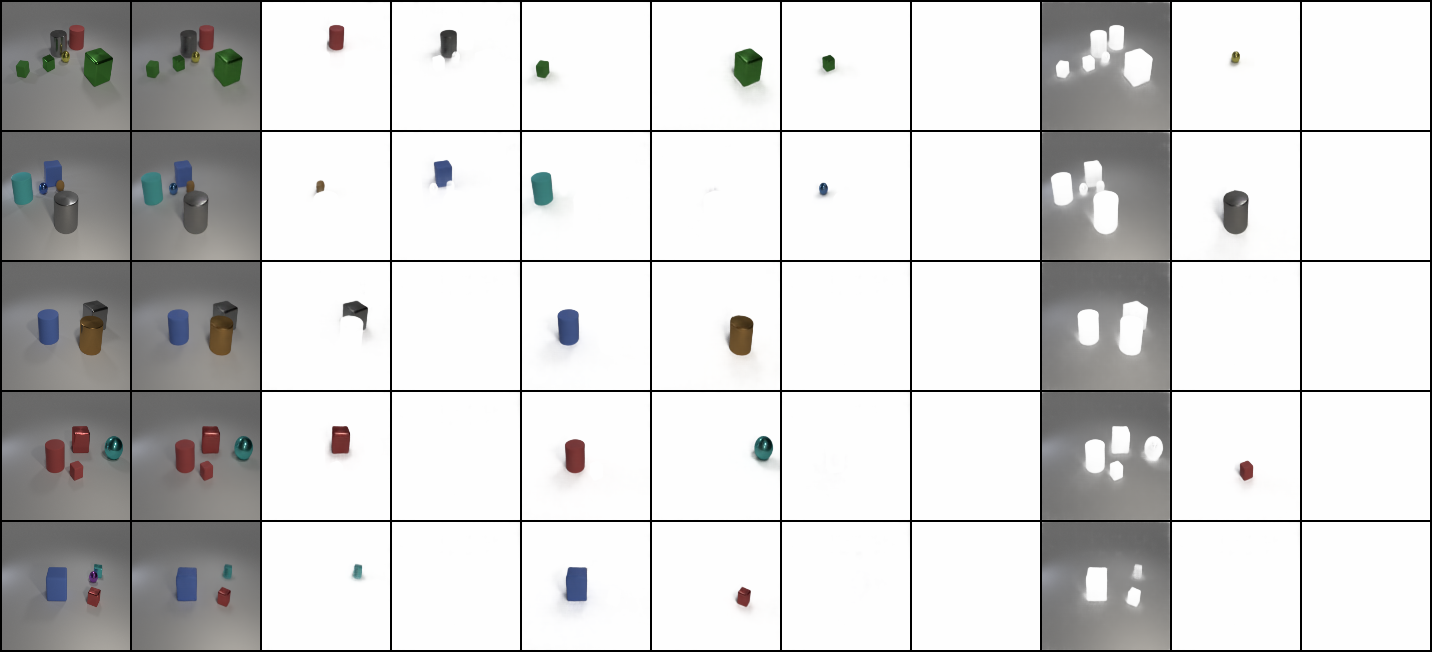}
    \caption{\textit{Results on CLEVR6:} The first column is the original image. The second column is the final reconstruction by the model, namely the weighted sum of individual reconstructions. Columns 3-11 are reconstructions of individual slots. The individual reconstructions are displayed without the mask.}
    \label{fig:clevr_result}
\end{figure}

\paragraph{Baselines}
We compare $\mathbf{C}$OP with the following commonly used slot extraction methods: Slot Attention \citep{SlotAttention}, IODINE \citep{IODINE} and MoNET \citep{Monet}.

\paragraph{Datasets}
For the experiments we use three standard benchmarks common in prior work, 
namely CLEVR6, Multi-dSprites and Tetrominoes \citep[introduced in][]{multiobjectdatasets19}.
These are synthetic datasets which provide ground truth segmentation masks for evaluation.
CLEVR consists of rendered scenes of 3D geometric shapes, e.g., spheres, cubes or cuboids, with a single light source.
CLEVR6 is a subset of CLEVR which only contains a maximum of six objects.
Multi-dSprites consists of 2-5 objects, which are 2D shapes (ellipses, hearts, squares) in different colors on black background.
Tetrominoes consists of always three objects, namely 2D shapes from the tetris game.

Occlusion is an important property that largely defines the difficulty of a dataset.
In tetrominoes there is no occlusion between objects.
For CLEVR6, objects may partially occlude each other, though usually a single object is not occluded by more than one other object.
In Multi-dSprites an object may be occluded by several other objects, which makes it the most challenging of the three tasks in this regard.   

For all these datasets we follow the protocol mentioned in Slot Attention \citep{SlotAttention}. 
CLEVR6 consists of 70k, and Multi-dSprites and Tetrominoes of 60k training samples. 
We partition the training set into a training and validation split to choose hyperparameters. For testing we trained on the whole training set and tested on 320 test samples (same as Slot Attention).

\paragraph{Training}
For the encoder and decoder architectures we follow the protocol described in \citep{SlotAttention}.
For the grouping module we use the architecture described in section \ref{sec:method} with slight modifications for each dataset.
First, we adjust the number of slots to equal or somewhat more than maximum number of objects of the according dataset. Since we have seen degrading performance of too many slots (also see section \ref{sec:ablations}). Specifically, for CLEVR6 and Multi-dSprites we use nine slots, and for Tetrominoes we use four slots (three object slots and one background slots).
All datasets are trained using a mean-squared-error reconstruction loss.
Full details about the hyperparameters can be found in the appendix \ref{app:hyperparameters}.

\paragraph{Evaluation metric}
All of the above-mentioned datasets contain ground truth masks that enable evaluation of the image segmentations. 
We use the Adjusted Rand Index \citep{RandIndex, Hubert1985_AdjustedRandIndex} as evaluation metric. 
This is a metric in the interval $[-0.5, 1.0]$ that compares two cluster assignments, whereas a score of $1$ means the clusters are identical, a score of $0$ would mean roughly equal to random assignment and $-0.5$ the worst assignment possible (worse than random).
As common in the literature \citep{SlotAttention, GenesisV2} we omit the background ground truth masks in the evaluation, since we mostly want to distinguish between foreground objects. This metric is referred to as the FG-ARI.

\begin{figure}
\centering
\begin{minipage}[b]{0.62\textwidth}
  \centering
    \includegraphics[width=\linewidth]{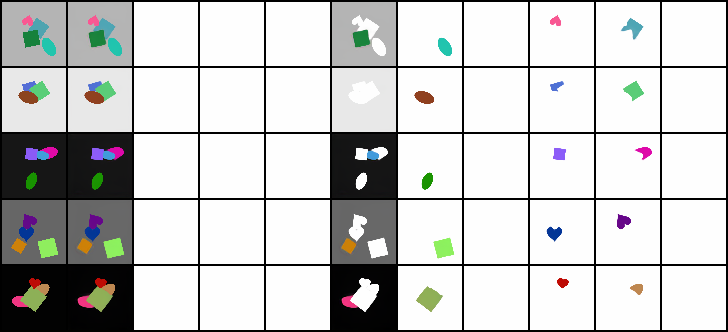}
\end{minipage}
\hfill
\begin{minipage}[b]{0.34\textwidth}
  \centering
    \includegraphics[width=\linewidth]{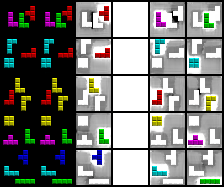}
\end{minipage}

\caption{
\textit{Left:} Reconstructions of Multi-dSprites. The first column is the original image. The second column is the weighted sum of individual reconstructions where the weights come from the masks to which a pixel-wise softmax was applied. Columns 3-11 are reconstructions of individual slots.
\textit{Right:} Reconstructions of Tetrominoes. Again, the first column is the original image, while the second column is the weighted sum of individual reconstructions. Columns 3-6 are reconstructions of individual slots.
The individual reconstructions are displayed without the mask.
}
\label{fig:mds_and_tetro_result}
\end{figure}

\begin{table}[b]
\caption{Results of different slot extraction methods on several datasets. Prior methods averaged over five seeds, while our results were averaged over three seeds.}
\label{tab:results}
\begin{center}
\begin{tabular}{lcccc}
\bf Method & \bf CLEVR6 & \bf Multi-dSprites & \bf Tetrominoes & \bf Average \\
\hline
Slot Attention & $\textbf{98.8} \pm 0.3$ & $91.3 \pm 0.3$ & $99.5 \pm 0.2 $ & \textbf{96.53} \\
IODINE & $\textbf{98.8} \pm 0.0$ & $76.7 \pm 5.6$ & $99.2 \pm 0.4 $ & 91.57 \\
MONET & $96.2 \pm 0.6$ & $90.4 \pm 0.8$ & n/a & 93.30 \\
\textbf{SAMP (ours)} & $97.6 \pm 0.6$ & $\textbf{92.3} \pm 0.2$ & $\textbf{99.8}\pm0.1$ & \textbf{96.57} \\
\end{tabular}
\end{center}
\end{table}

\paragraph{Results}

SAMP is competitive for CLEVR6, but outperforms Slot Attention in Multi-dSprites and Tetrominoes.
SAMP achieves state-of-the-art performance --- on par with Slot Attention --- when averaged over all three datasets.
An overview of the results is given in table \ref{tab:results}.
Qualitatively we can see in figures \ref{fig:mds_and_tetro_result} and \ref{fig:clevr_result} that the slot reconstructions are semantically meaningful and of high quality.
Interestingly, some slots seem to specialize.
E.g., for CLEVR6 and Tetrominoes there seems to be a clear specialization of the second slot to capture the gray background. 
This is also consistent with our observations of the slots over the training process. Non-background slots don't seem to specialize in the position, shape, size nor color of the objects.
Interestingly, some slots are not used in \emph{any} of the reconstructions, which can be regarded as a specialization on "not interfering" with other reconstructions.


\begin{figure}[t]
    \centering
    \includegraphics[width=\linewidth]{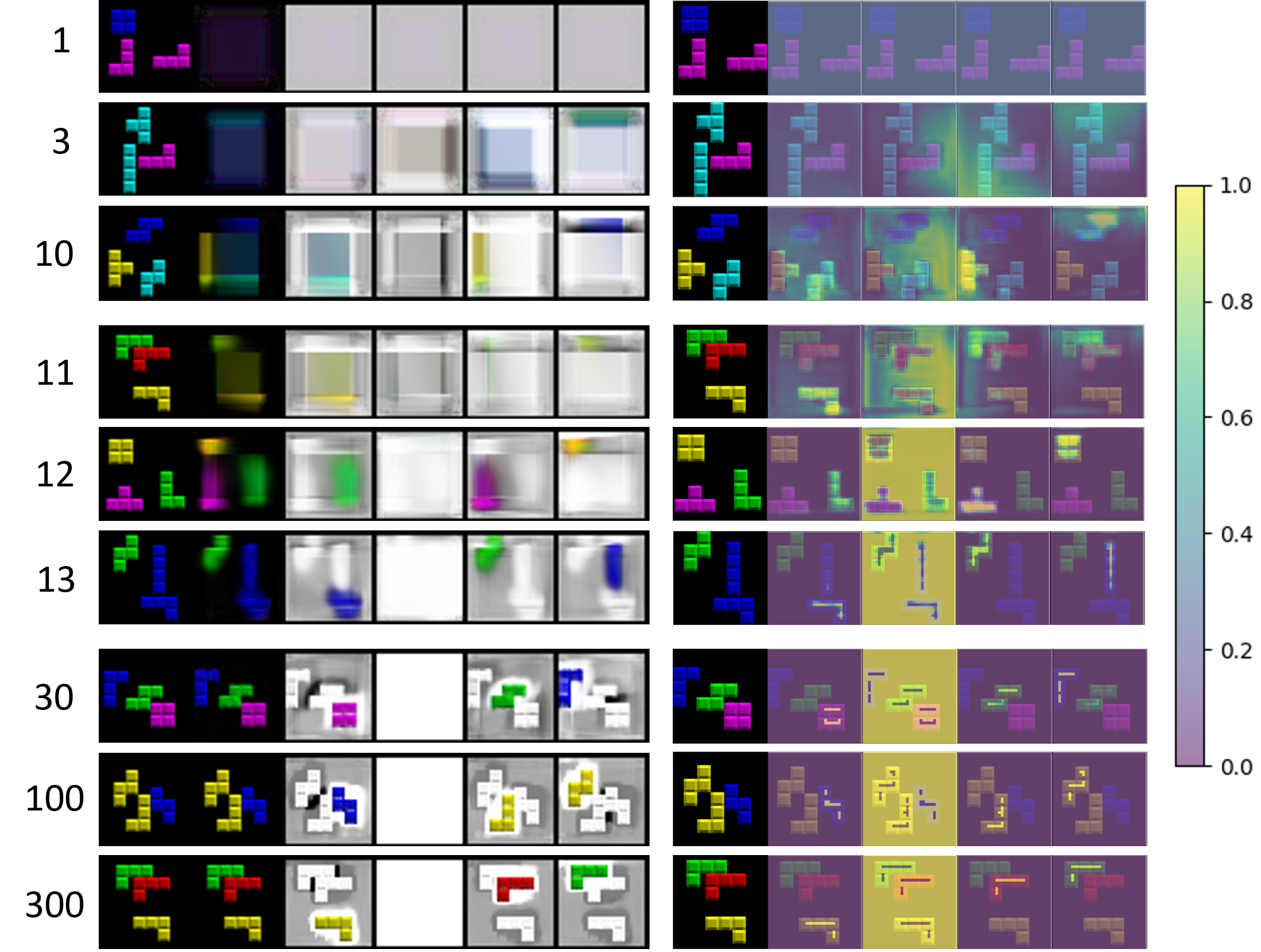}
    \caption{\textit{Reconstructions and visualized attention heatmaps of slots over pixel features during training on Tetrominoes:} The numbers on the left denote the completed training epochs, the left group of images are reconstructions, whereas the right group are visualized attention maps.
    The columns of the reconstruction images are in the following order: (col. 1) ground truth image, (col. 2) final reconstruction (cols. 3-6) individual slot reconstructions. 
    The columns of the visualized attention maps are in the following order: (col.1) ground truth image, (cols. 2-5) individual attention maps of the queries over the keys (i.e. the projected pixel features).
    }
    \label{fig:tetro_visualizations}
\end{figure}

\section{Analysis and Ablation studies}
\label{sec:ablations}

\paragraph{Learned query visualization}
We visualized the attention of a slot over the keys of the pixel features.
The results on the Tetrominoes dataset are in figure \ref{fig:tetro_visualizations} (for further visualizations, please see appendix \ref{app:ablations}).
For the Tetrominoes run in figure \ref{fig:tetro_visualizations}, we can roughly identify three phases. 
\begin{itemize}
    \item \textbf{Warmup phase} Up until epoch 10, the reconstructions are poor and the attention maps are not sharp nor object-specific.
    \item \textbf{Transition phase} From epoch 11 to 13, there is a transition phase. While reconstructions are still poor after epoch 11, they have become sharper and object-specific after epoch 13. Also the attention maps clearly indicate that the queries are focusing on object-specific parts of the input.
    \item \textbf{Refinement phase} The objects at epoch 30 are still somewhat blurry, but are more detailed refined over the next several hundred epochs.
    For the attention maps, we can see that between epoch 13 and epoch 30 the attention maps change little. From epoch 30 onwards, the queries attend to very similar pixels of individual objects.
\end{itemize}

\paragraph{Which attention mechanism to use}
To understand the impact of the attention mechanism, we tried several ablations.
\begin{itemize}
    \item \textbf{SAMP w/o SSA Layer} To check if SSA Layer over pixel features is beneficial at all, we omit the attention mechanism and directly reconstruct the images from the queries.
    \item \textbf{SAMP w/ Cross-Attention} Furthermore we substitute the SSA Layer with the common Cross-Attention \citep{Transformer_Vaswani}.
    \item \textbf{SAMP w/ Attention as in Slot Attention} we substitute the SSA Layer with the attention variant used in Slot Attention \citep{SlotAttention}. Note that the GRU, MLPs at the end of each iteration were not used. Also we just applied the attention-variant once in a non-iterative fashion.
\end{itemize}
The results can be found in table \ref{table:ablations}.
With four slots not using an attention-variant leads to a substantial drop in the FG-ARI score by 1.8.
We conclude that competition of slots over pixel features is a useful bias for this architecture.

\paragraph{The effect of the maximum number of slots}
To gain further insights on how SAMP generalizes to more slots, we ran an ablation on Tetrominoes (Table \ref{table:ablations}).
While SAMP manages to efficiently solve the task for six slots --- which are 50\% more than the ideal number slots --- the performance largely degrades at nine or more slots. As visible in figure \ref{fig:ablation_9_slots}, oftentimes information about a single object is then shared among two slots.

\begin{table}[ht]
    \caption{
    FG-ARI Mean and standard deviations over three seeds on the Tetrominoes dataset.
    \textit{Left:} Ablations on the number of maximum slots.
    The ideal number of slots is four (i.e., three foreground objects and the background). If the number of slots is six (i.e., 50\% higher than the ideal number of slots), the model still learns reasonably. If the number of slots becomes twice the number of ideal slots, the models performance degrades rapidly.
    \textit{Right: } A comparison of attention-variants. The currently chosen attention-variant has the strongest performance. Using no attention mechanism (i.e., using the queries directly for reconstruction), Cross-Attention or the attention mechanism used in the Slot Attention module degrade performance.
    }
    \label{table:ablations}
    \centering
    \begin{minipage}{.3\textwidth}
        \begin{center}
        \begin{tabular}{ll}
        \multicolumn{1}{c}{\bf Configuration} & \multicolumn{1}{c}{\bf Tetrominoes} \\
        \hline
        \textbf{SAMP (4 slots)} & $\mathbf{99.77 \pm 0.12}$ \\
        SAMP (6 slots) & $99.58 \pm 0.65$ \\
        SAMP (8 slots) & $89.13 \pm 18.58$ \\
        SAMP (9 slots) & $84.21 \pm 13.62$ \\
        SAMP (16 slots) & $63.99 \pm 1.57$ \\
        \end{tabular}
        \end{center}
    \end{minipage}%
    \hspace{\dimexpr 0.1\textwidth}
    \begin{minipage}{.5\textwidth}
        \begin{center}
        \begin{tabular}{ll}
        \multicolumn{1}{c}{\bf Attention variants (4 slots)} & \multicolumn{1}{c}{\bf Tetrominoes} \\
        \hline
        \textbf{SAMP} & $\mathbf{99.77 \pm 0.12}$ \\
        SAMP w/o Attention & $97.97 \pm 0.20$ \\
        SAMP w/ Attn as in Slot Attention & $91.07 \pm 15.23$ \\
        SAMP w/ Cross-Attention & $90.92 \pm 14.87$ \\
        \\
        \end{tabular}
        \end{center}
    \end{minipage}
\end{table}

\section{Limitations and Discussion}

\paragraph{Sensitivity to the maximum number of slots}
One down-side of SAMP is, that the number of objects is not adjustable during or after training.
Other methods also suffer from a sensitivity to the number of slots during training. This problem can be improved by training with more slots and flexibly adjust the number of slots during inference \citep{zimmermann2023_SlotSensitivity}.


\paragraph{Connections to continual learning}
The ability to learn abstract, object-centric representations in an unsupervised manner is a key challenge for
continual learning agents that must operate in dynamic, continually changing environments \citep{greff-Binding_problem, wang2024LifeLongLearningReview}.
By extracting slots that correspond to object concepts from raw sensory data, SAMP provides a mechanism for continually building up a structured understanding of the world in a continual learning setting.
As an agent experiences more data over time, the slots can be refined and updated to reflect new objects and concepts encountered (e.g. similar to \cite{seitzer2023dinosaur}). 
This allows for efficient knowledge accumulation and transfer as the agent faces distribution shifts and must adapt to new situations, a core goal of continual learning. 
Furthermore, our simple, scalable architecture based on standard primitives like convolutions and attention makes SAMP well-suited for continual learning scenarios with constrained computational resources.  Overall, unsupervised object-centric representation learning enabled by approaches like SAMP can serve as a fundamental building block for continual learning agents.

\section{Conclusion}

In this paper we presented a novel method for object-centric learning that is based on competition, called Simplified Slot Attention with Max Pool Priors (SAMP).
SAMP is simple, scalable, fully-differentiable and contrary to the widely applied Slot Attention, it is non-iterative.
It comes with the down-side of not being able to adjust to a different number of slots between training and test time.
Empirically our method is competitive or outperforms existing slot extraction methods.

\section*{Acknowledgements}
The ELLIS Unit Linz, the LIT AI Lab, and the Institute for Machine Learning are supported by the Federal State of Upper Austria. 
We thank the projects Medical Cognitive Computing Center (MC3), INCONTROL-RL (FFG-881064), PRIMAL (FFG-873979), S3AI (FFG-872172), EPILEPSIA (FFG-892171), AIRI FG 9-N (FWF-36284, FWF-36235), AI4GreenHeatingGrids (FFG- 899943), INTEGRATE (FFG-892418), ELISE (H2020-ICT-2019-3 ID: 951847), Stars4Waters (HORIZON-CL6-2021-CLIMATE-01-01).
We thank the European High Performance Computing initiative for providing computational resources (EHPC-DEV-2023D08-019).
We thank Audi.JKU Deep Learning Center, TGW LOGISTICS GROUP GMBH, Silicon Austria Labs (SAL), FILL Gesellschaft mbH, Anyline GmbH, Google, ZF Friedrichshafen AG, Robert Bosch GmbH, UCB Biopharma SRL, Merck Healthcare KGaA, Verbund AG, Software Competence Center Hagenberg GmbH, Borealis AG, T\"{U}V Austria, Frauscher Sensonic, TRUMPF, and the NVIDIA Corporation.

\bibliography{collas2024_conference}
\bibliographystyle{collas2024_conference}

\appendix
\section{Appendix}

\subsection{Ablations}
\label{app:ablations}

\begin{figure}[h]
    \centering
    \includegraphics[width=\linewidth]{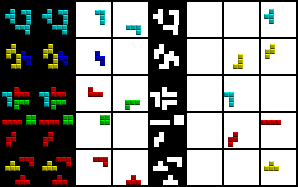}
    \caption{\textit{Ablation of Tetrominoes with six slots:} The reconstructions still work reasonably well.}
    \label{fig:ablation_6_slots}
\end{figure}

\begin{figure}[h]
    \centering
    \includegraphics[width=\linewidth]{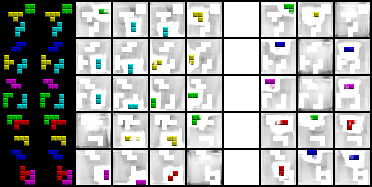}
    \caption{\textit{Ablation of Tetrominoes with 8 slots:} Since there are twice as many slots than the ideal four for Tetrominoes (i.e., three objects, one background), a single object is often represented in two slots.}
    \label{fig:ablation_8_slots}
\end{figure}

\begin{figure}[h]
    \centering
    \includegraphics[width=\linewidth]{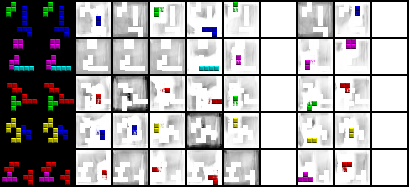}
    \caption{\textit{Ablation of Tetrominoes with nine slots:} Since there are more than twice as many slots than the ideal four for Tetrominoes (i.e., three objects, one background), a single object is often represented in two slots.}
    \label{fig:ablation_9_slots}
\end{figure}

In this section we add additional figures for the ablation studies. The figures are taken from \textbf{the best prediction of the worst run} among all the seeds.
Figures \ref{fig:ablation_6_slots}, \ref{fig:ablation_8_slots} and \ref{fig:ablation_9_slots} are reconstructions of the respective architectures.

\subsection{Hyperparameters}
\label{app:hyperparameters}

\subsubsection{Encoder architectures}
For the encoder architectures we follow the protocol given in \citet{SlotAttention}.
The detailed layer descriptions for CLEVR6 can be found in table \ref{tab:encoder_clevr}. 
Similarly the encoder layer descriptions for Multi-dSprites and Tetrominoes are explained in table \ref{tab:encoder_mds_tetro}.

\begin{table}[h]
\caption{CNN encoder for CLEVR6.}
\label{tab:encoder_clevr}
\begin{center}
\begin{tabular}{cccc}
\textbf{Type} & \textbf{Size/Channels} & \textbf{Activation} & \textbf{Comment} \\
\hline
Conv 5 $\times$ 5 & 64 & ReLU & Stride: 1 \\
Conv 5 $\times$ 5 & 64 & ReLU & Stride: 1 \\
Conv 5 $\times$ 5 & 64 & ReLU & Stride: 1 \\
Conv 5 $\times$ 5 & 64 & ReLU & Stride: 1 \\
Position Embedding & - & - & See Section \ref{app:pos_emb} \\
Flatten & axis: [0, 1 $\times$ 2, 3] & - & Flatten x, y pos. \\
Layer Norm & - & - & - \\
MLP (per location) & 64 & ReLU & - \\
MLP (per location) & 64 & - & - \\
\end{tabular}
\end{center}
\end{table}

\begin{table}[h]
\caption{CNN encoder for Multi-dSprites and Tetrominoes.}
\label{tab:encoder_mds_tetro}
\begin{center}
\begin{tabular}{cccc}
\textbf{Type} & \textbf{Size/Channels} & \textbf{Activation} & \textbf{Comment} \\
\hline
Conv 5 $\times$ 5 & 32 & ReLU & Stride: 1 \\
Conv 5 $\times$ 5 & 32 & ReLU & Stride: 1 \\
Conv 5 $\times$ 5 & 32 & ReLU & Stride: 1 \\
Conv 5 $\times$ 5 & 32 & ReLU & Stride: 1 \\
Position Embedding & - & - & See Section \ref{app:pos_emb} \\
Flatten & axis: [0, 1 $\times$ 2, 3] & - & Flatten x, y pos. \\
Layer Norm & - & - & - \\
MLP (per location) & 32 & ReLU & - \\
MLP (per location) & 32 & - & - \\
\end{tabular}
\end{center}
\end{table}

\subsubsection{Competition encoder architectures}

The layer descriptions for the competition encoders can be found in  tables \ref{tab:comp_enc_tetro}, \ref{tab:comp_enc_clevr} and \ref{tab:comp_enc_mds}.
Note that before the Competition encoder, a positional bias is added to the pixel features, as described in appendix \ref{app:pos_emb}.

\begin{table}[h]
\caption{Competition encoder layers for Tetrominoes.}
\label{tab:comp_enc_tetro}
\begin{center}
\begin{tabular}{lccccc}
\textbf{Type} & \textbf{Spatial Resolution} & \textbf{Size/Channels} & \textbf{Activation} & \textbf{Stride} & \textbf{Padding} \\
\hline
Conv 5 $\times$ 5 & 35x35 & 32 & LeakyReLU & 1 & 2 \\
MaxPool & 17x17 & 32 & & 2 & 0 \\
Conv 5 $\times$ 5 & 17x17 & 32 & LeakyReLU & 1 & 2 \\
MaxPool & 8x8 & 32 & & 2 & 0 \\
Conv 5 $\times$ 5 & 8x8 & 32 & LeakyReLU & 1 & 2 \\
MaxPool & 4x4 & 32 & & 2 & 0 \\
Conv 5 $\times$ 5 & 4x4 & 32 & LeakyReLU & 1 & 2 \\
MaxPool & 2x2 & 32 & & 2 & 0 \\
Flatten & 2x2 $\rightarrow$ 4 & axis: [0, 1 $\times$ 2, 3] & & & \\
FC & 4 & 32 & LeakyReLU & & \\
FC & 4 & 32 & & & \\
\end{tabular}
\end{center}
\end{table}

\begin{table}[h]
\caption{Competition encoder layers for Multi-dSprites.}
\label{tab:comp_enc_mds}
\begin{center}
\begin{tabular}{lccccc}
\textbf{Type} & \textbf{Spatial Resolution} & \textbf{Size/Channels} & \textbf{Activation} & \textbf{Stride} & \textbf{Padding} \\
\hline
Conv 5 × 5 & 64x64 & 64 & LeakyReLU & 1 & 2 \\
MaxPool & 64x64 & 64 & & 2 & 0 \\
Conv 5 × 5 & 32x32 & 64 & LeakyReLU & 1 & 2 \\
MaxPool & 32x32 & 64 & & 2 & 0 \\
Conv 5 × 5 & 16x16 & 64 & LeakyReLU & 1 & 2 \\
MaxPool & 16x16 & 64 & & 2 & 0 \\
Conv 5 × 5 & 8x8 & 64 & LeakyReLU & 1 & 2 \\
MaxPool & 8x8 & 64 & & 2 & 0 \\
Conv 5 × 5 & 4x4 & 64 & LeakyReLU & 1 & 2 \\
MaxPool & 4x4 & 64 & & 1 & 0 \\
Flatten & 3x3 $\rightarrow$ 9 & axis: [0, 1 × 2, 3] & & & \\
FC & 9 & 64 & LeakyReLU & & \\
FC & 9 & 64 & & & \\
\end{tabular}
\end{center}
\end{table}

\begin{table}[h]
\caption{Competition encoder layers for CLEVR6.}
\label{tab:comp_enc_clevr}
\begin{center}
\begin{tabular}{lccccc}
\textbf{Type} & \textbf{Spatial Resolution} & \textbf{Size/Channels} & \textbf{Activation} & \textbf{Stride} & \textbf{Padding} \\
\hline
Conv 5 × 5 & 128x128 & 64 & LeakyReLU & 1 & 2 \\
MaxPool & 128x128 & 64 & & 2 & 0 \\
Conv 5 × 5 & 64x64 & 64 & LeakyReLU & 1 & 2 \\
MaxPool & 64x64 & 64 & & 2 & 0 \\
Conv 5 × 5 & 32x32 & 64 & LeakyReLU & 1 & 2 \\
MaxPool & 32x32 & 64 & & 2 & 0 \\
Conv 5 × 5 & 16x16 & 64 & LeakyReLU & 1 & 2 \\
MaxPool & 16x16 & 64 & & 2 & 0 \\
Conv 5 × 5 & 8x8 & 64 & LeakyReLU & 1 & 2 \\
MaxPool & 8x8 & 64 & & 2 & 0 \\
Conv 5 × 5 & 4x4 & 64 & LeakyReLU & 1 & 2 \\
MaxPool & 4x4 & 64 & & 1 & 0 \\
Flatten & 3x3 $\rightarrow$ 9 & axis: [0, 1 × 2, 3] & & & \\
FC & 9 & 64 & LeakyReLU & & \\
FC & 9 & 64 & & & \\
\end{tabular}
\end{center}
\end{table}

\subsubsection{Decoder architectures}
Again, we follow the protocol of \citet{SlotAttention}.
Note that before the decoding, a positional bias is added to the spatial map, as described in appendix \ref{app:pos_emb}.
The detailed layer descriptions for CLEVR6 can be found in table \ref{tab:decoder_clevr}. 
Similarly the encoder layer descriptions for Multi-dSprites and Tetrominoes are explained in table \ref{tab:decoder_mds_tetro}.

\begin{table}[h]
\caption{Decoder layer description for CLEVR6.}
\label{tab:decoder_clevr}
\begin{center}
\begin{tabular}{lcccl}
\textbf{Type} & \textbf{Spatial res.} & \textbf{Size/Channels} & \textbf{Activation} & \textbf{Comment} \\
\hline
Spatial Broadcast & 1 & 64 & - & - \\
Position Embedding & 8x8 & 64 & - & See \ref{app:pos_emb} \\
Conv 5 x 5 & 8x8 & 64 & ReLU & Stride: 2 \\
Conv 5 x 5 & 16x16 & 64 & ReLU & Stride: 2 \\
Conv 5 x 5 & 32x32 & 64 & ReLU & Stride: 2 \\
Conv 5 x 5 & 64x64 & 64 & ReLU & Stride: 2 \\
Conv 5 x 5 & 128x128 & 64 & ReLU & Stride: 1 \\
Conv 3 x 3 & 128x128 & 4 & - & Stride: 1 \\
Split Channels & 128x128 & RGB (3), mask (1) & Softmax on masks & - \\
Recombine Slots & 128x128 & - & - & - \\
\end{tabular}
\end{center}
\end{table}

\begin{table}[h]
\caption{Decoder layer description for Multi-dSprites and Tetrominoes.}
\label{tab:decoder_mds_tetro}
\centering
\begin{tabular}{lccccl}
\hline
\textbf{Type} & \textbf{Spatial res.} & \textbf{Size/Channels} & \textbf{Activation} & \textbf{Comment} \\
\hline
Spatial Broadcast & 1 & 32 & - & - \\
Position Embedding & WxH & 32 & - & See \ref{app:pos_emb} \\
Conv 5 x 5 & WxH & 32 & ReLU & Stride: 1 \\
Conv 5 x 5 & WxH & 32 & ReLU & Stride: 1 \\
Conv 5 x 5 & WxH & 32 & ReLU & Stride: 1 \\
Conv 3 x 3 & WxH & 4 & - & Stride: 1 \\
Split Channels & WxH & RGB (3), mask (1) & Softmax on masks & - \\
Recombine Slots & WxH & - & - & - \\
\end{tabular}
\end{table}

\subsubsection{Positional Embedding}
\label{app:pos_emb}

We follow the positional embedding given in \citet{SlotAttention}.
Summarized, for each pixel a 4-dimensional positional embedding is initialized that encodes the distance to all colors (normalized to the range $[0, 1]$). These 4 dimensional embeddings are then projected with a learnable matrix to the dimensionality of the pixel features. 

\subsection{Learned query visualizations}
\label{app:learned_query_vis}

\begin{figure}[h]
    \centering
    \includegraphics[width=\linewidth]{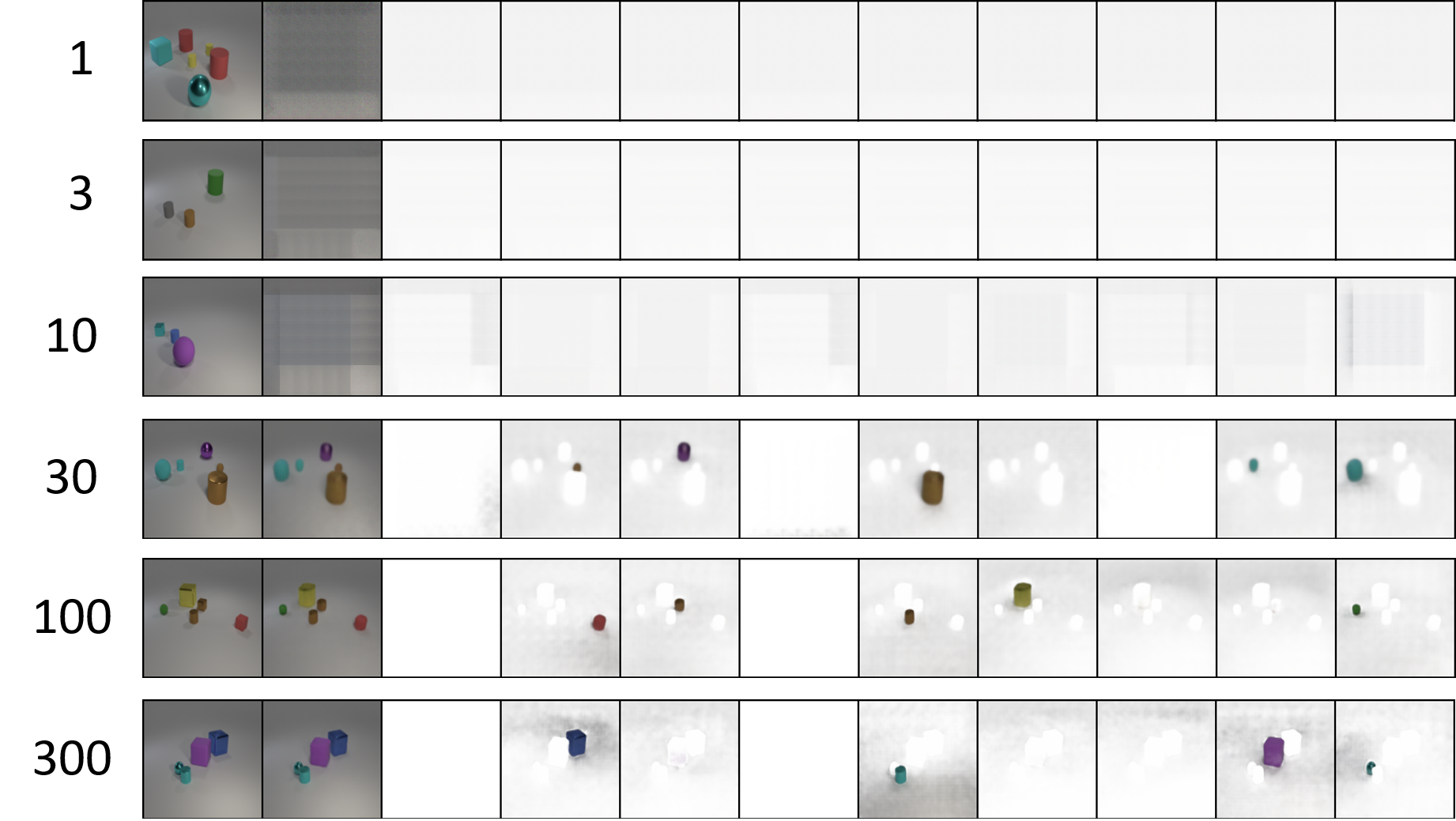}
    \caption{\textit{Reconstructions over training of CLEVR6}:
    The numbers on the left denote the completed training epochs, whereas the images are reconstructions.
    The columns of the reconstruction images are in the following order: (col. 1) ground truth image, (col. 2) reconstruction (cols. 3-11) individual slot reconstructions. 
    }
    \label{fig:reconstructions_clevr}
\end{figure}

\begin{figure}[h]
    \centering
    \includegraphics[width=\linewidth]{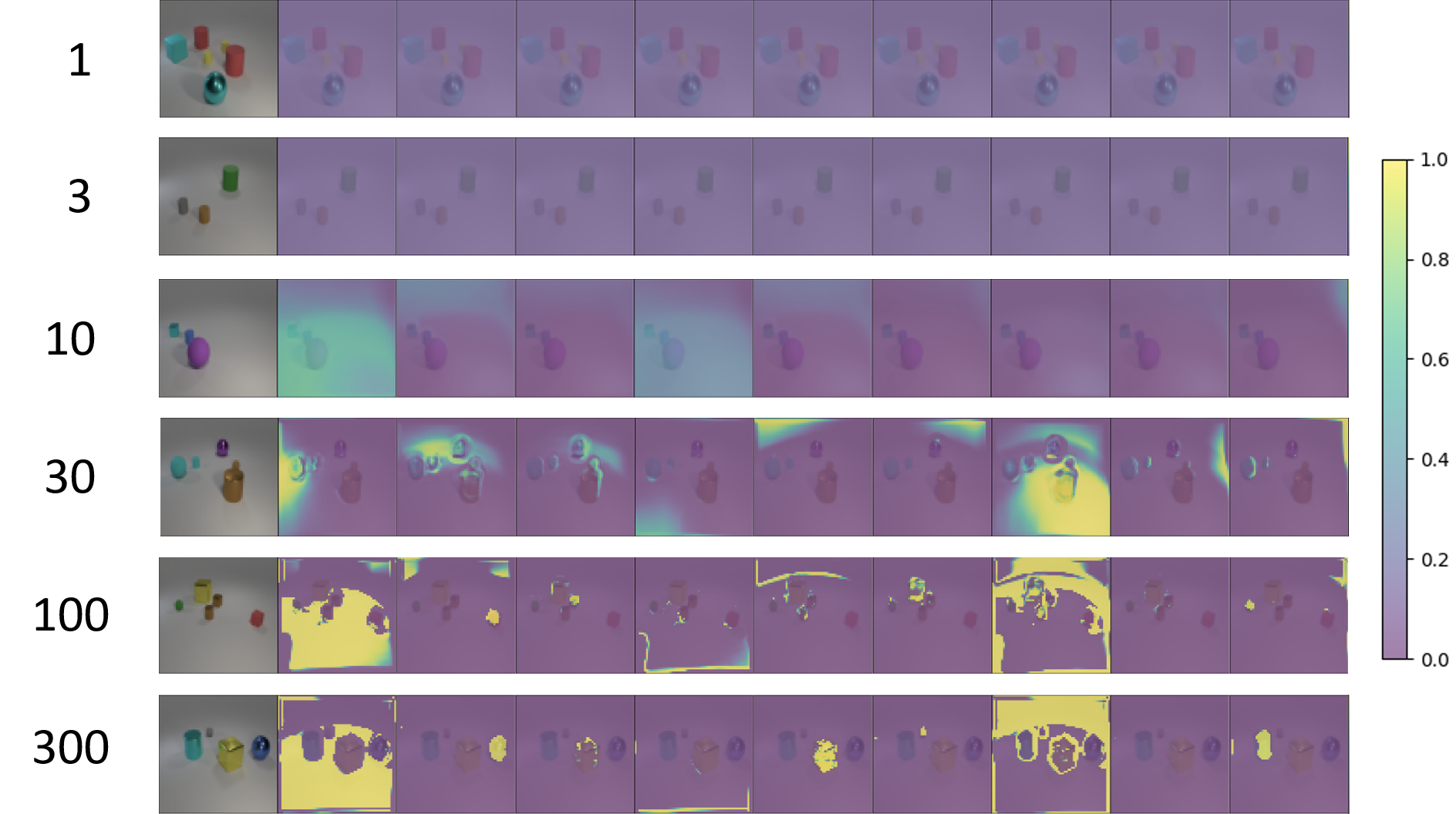}
    \caption{\textit{Attention heatmaps during training of CLEVR6}:
    The numbers on the left denote the completed training epochs, whereas the images are visualized attention maps. 
    The columns of the visualized attention maps are in the following order: (col.1) ground truth image, (cols. 2-10) individual attention maps of the queries over the keys (transformed pixel features).
    }
    \label{fig:heatmaps_clevr}
\end{figure}

Similarly to the Tetrominoes, we see that the recognition of objects gradually evolves for CLEVR6 and Multi-dSprites.
The learning process over time of CLEVR6 is depicted in figure \ref{fig:reconstructions_clevr} for the reconstructions and in figure \ref{fig:heatmaps_clevr} for the attention heatmaps.

\begin{figure}[h]
    \centering
    \includegraphics[width=\linewidth]{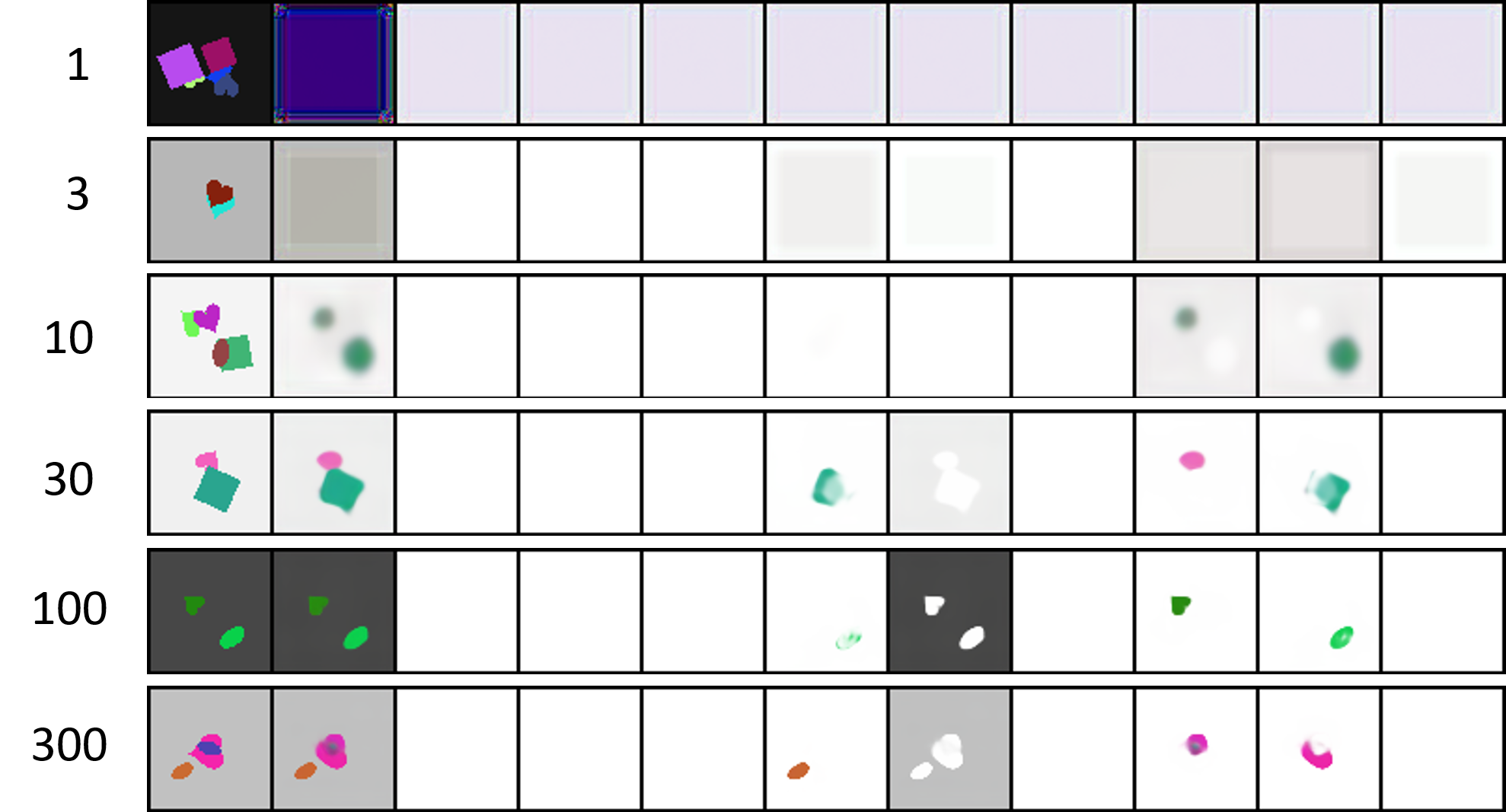}
    \caption{\textit{Reconstructions over training of Multi-dSprites}:
    The numbers on the left denote the completed training epochs, whereas the images are reconstructions.
    The columns of the reconstruction images are in the following order: (col. 1) ground truth image, (col. 2) reconstruction (cols. 3-11) individual slot reconstructions. 
    }
    \label{fig:reconstructions_sprites}
\end{figure}

\begin{figure}[h]
    \centering
    \includegraphics[width=\linewidth]{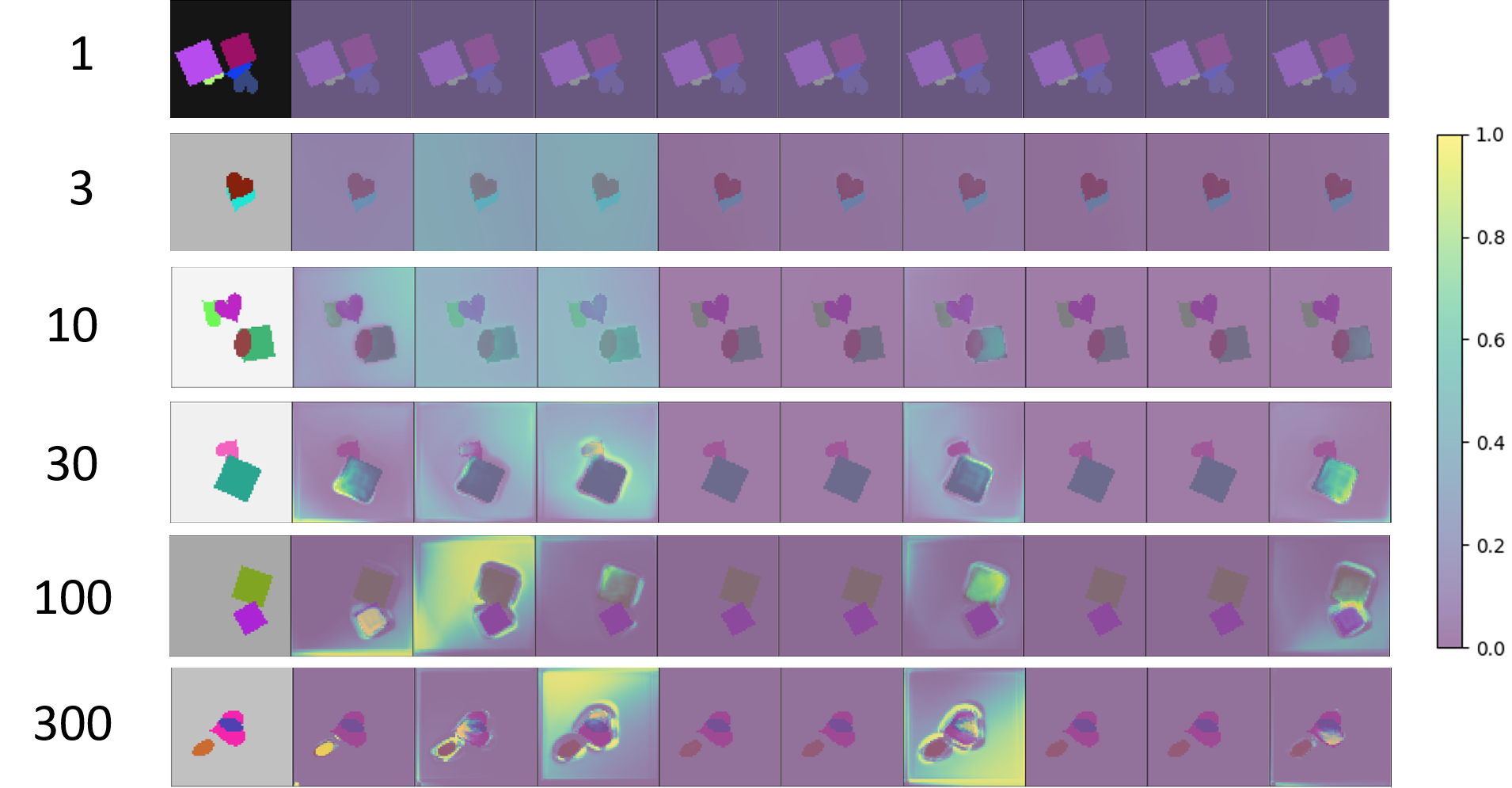}
    \caption{\textit{Heatmaps of attention over training of Multi-dSprites}:
    The numbers on the left denote the completed training epochs, whereas the images are visualized attention maps. 
    The columns of the visualized attention maps are in the following order: (col.1) ground truth image, (cols. 2-10) individual attention maps of the queries over the keys (transformed pixel features).
    }
    \label{fig:heatmaps_sprites}
\end{figure}

The learning process over time of Multi-dSprites is depicted in figure \ref{fig:reconstructions_sprites} for the reconstructions and figure \ref{fig:heatmaps_sprites} for the attention heatmaps.
Although some slots do not show a visible attention map, we checked their numerical value and conclude that the attention values are chosen large enough to reconstruct the objects.

\end{document}

%% file: math_commands.tex
\usepackage{amsmath,amsfonts,bm}

\def\eqref#1{equation~\ref{#1}}

\def\1{\bm{1}}

\DeclareMathAlphabet{\mathsfit}{\encodingdefault}{\sfdefault}{m}{sl}
\SetMathAlphabet{\mathsfit}{bold}{\encodingdefault}{\sfdefault}{bx}{n}

\newcommand{\R}{\mathbb{R}}



%% file: collas2024_conference.bbl
\begin{thebibliography}{62}
\providecommand{\natexlab}[1]{#1}
\providecommand{\url}[1]{\texttt{#1}}
\expandafter\ifx\csname urlstyle\endcsname\relax
  \providecommand{\doi}[1]{doi: #1}\else
  \providecommand{\doi}{doi: \begingroup \urlstyle{rm}\Url}\fi

\bibitem[Andersen et~al.(1969)Andersen, Gross, Lomo, and Sveen]{Andersen:69}
Per Andersen, G.~N. Gross, Torkel Lomo, and Ole Sveen.
\newblock Participation of inhibitory and excitatory interneurones in the control of hippocampal cortical output.
\newblock \emph{UCLA Forum Med Sci}, 11:\penalty0 415--465, 1969.

\bibitem[Burgess et~al.(2019)Burgess, Matthey, Watters, Kabra, Higgins, Botvinick, and Lerchner]{Monet}
Christopher~P. Burgess, Loic Matthey, Nicholas Watters, Rishabh Kabra, Irina Higgins, Matt Botvinick, and Alexander Lerchner.
\newblock Monet: Unsupervised scene decomposition and representation, 2019.

\bibitem[Chang et~al.(2023)Chang, Griffiths, and Levine]{ImplicitSlotAttention}
Michael Chang, Thomas~L. Griffiths, and Sergey Levine.
\newblock Object representations as fixed points: Training iterative refinement algorithms with implicit differentiation, 2023.

\bibitem[Chung et~al.(2014)Chung, G{\"{u}}l{\c{c}}ehre, Cho, and Bengio]{ChungGCB14_GRU}
Junyoung Chung, {\c{C}}aglar G{\"{u}}l{\c{c}}ehre, KyungHyun Cho, and Yoshua Bengio.
\newblock Empirical evaluation of gated recurrent neural networks on sequence modeling.
\newblock \emph{CoRR}, abs/1412.3555, 2014.
\newblock URL \url{http://arxiv.org/abs/1412.3555}.

\bibitem[Comaniciu \& Meer(2002)Comaniciu and Meer]{Mean_Shift_Clustering}
D.~Comaniciu and P.~Meer.
\newblock Mean shift: a robust approach toward feature space analysis.
\newblock \emph{IEEE Transactions on Pattern Analysis and Machine Intelligence}, 24\penalty0 (5):\penalty0 603--619, 2002.
\newblock \doi{10.1109/34.1000236}.

\bibitem[Deng et~al.(2009)Deng, Dong, Socher, Li, Li, and Fei-Fei]{Deng:09}
Jia Deng, Wei Dong, Richard Socher, Li-Jia Li, Kai Li, and Li~Fei-Fei.
\newblock Imagenet: A large-scale hierarchical image database.
\newblock pp.\  248--255, 2009.
\newblock \doi{10.1109/CVPR.2009.5206848}.

\bibitem[Eccles et~al.(1967)Eccles, Ito, and Szentágothai]{Eccles:67}
John~C. Eccles, Masao Ito, and János Szentágothai.
\newblock \emph{The Cerebellum as a Neuronal Machine}.
\newblock 1967.

\bibitem[Ellias \& Grossberg(1975)Ellias and Grossberg]{Ellias:75}
Stephen~A. Ellias and Stephen Grossberg.
\newblock Pattern formation, contrast control, and oscillations in the short term memory of shunting on-center off-surround networks.
\newblock \emph{Biol. Cybernetics}, 20:\penalty0 69--98, 1975.
\newblock \doi{10.1007/BF00327046}.
\newblock URL \url{https://doi.org/10.1007/BF00327046}.

\bibitem[Engelcke et~al.(2019)Engelcke, Kosiorek, Jones, and Posner]{Genesis}
Martin Engelcke, Adam~R. Kosiorek, Oiwi~Parker Jones, and Ingmar Posner.
\newblock {GENESIS:} generative scene inference and sampling with object-centric latent representations.
\newblock \emph{CoRR}, abs/1907.13052, 2019.
\newblock URL \url{http://arxiv.org/abs/1907.13052}.

\bibitem[Engelcke et~al.(2020)Engelcke, Jones, and Posner]{engelcke2020reconstruction}
Martin Engelcke, Oiwi~Parker Jones, and Ingmar Posner.
\newblock Reconstruction bottlenecks in object-centric generative models, 2020.

\bibitem[Engelcke et~al.(2021)Engelcke, Jones, and Posner]{GenesisV2}
Martin Engelcke, Oiwi~Parker Jones, and Ingmar Posner.
\newblock {GENESIS-V2:} inferring unordered object representations without iterative refinement.
\newblock \emph{CoRR}, abs/2104.09958, 2021.
\newblock URL \url{https://arxiv.org/abs/2104.09958}.

\bibitem[Ermentrout(1992)]{Ermentrout:92}
Bard Ermentrout.
\newblock Complex dynamics in winner-take-all neural nets with slow inhibition.
\newblock \emph{Neural Networks}, 5\penalty0 (3):\penalty0 415--431, 1992.
\newblock ISSN 0893-6080.
\newblock \doi{https://doi.org/10.1016/0893-6080(92)90004-3}.
\newblock URL \url{https://www.sciencedirect.com/science/article/pii/0893608092900043}.

\bibitem[Gao et~al.(2023)Gao, Hohmann, and Neumann]{gao2023_SA_with_cluster_init}
Ning Gao, Bernard Hohmann, and Gerhard Neumann.
\newblock Enhancing interpretable object abstraction via clustering-based slot initialization, 2023.

\bibitem[Givan et~al.(2003)Givan, Dean, and Greig]{Givan:03}
R.~Givan, T.~Dean, and M.~Greig.
\newblock Equivalence notions and model minimization in {Markov} decision processes.
\newblock \emph{Artificial Intelligence}, 147\penalty0 (1):\penalty0 163--223, 2003.
\newblock \doi{10.1016/S0004-3702(02)00376-4}.

\bibitem[Greff et~al.(2019)Greff, Kaufman, Kabra, Watters, Burgess, Zoran, Matthey, Botvinick, and Lerchner]{IODINE}
Klaus Greff, Rapha{\"{e}}l~Lopez Kaufman, Rishabh Kabra, Nick Watters, Chris Burgess, Daniel Zoran, Lo{\"{\i}}c Matthey, Matthew~M. Botvinick, and Alexander Lerchner.
\newblock Multi-object representation learning with iterative variational inference.
\newblock \emph{CoRR}, abs/1903.00450, 2019.
\newblock URL \url{http://arxiv.org/abs/1903.00450}.

\bibitem[Greff et~al.(2020)Greff, van Steenkiste, and Schmidhuber]{greff-Binding_problem}
Klaus Greff, Sjoerd van Steenkiste, and J{\"{u}}rgen Schmidhuber.
\newblock On the binding problem in artificial neural networks.
\newblock \emph{CoRR}, abs/2012.05208, 2020.
\newblock URL \url{https://arxiv.org/abs/2012.05208}.

\bibitem[He et~al.(2016)He, Zhang, Ren, and Sun]{He:16}
K.~He, X.~Zhang, S.~Ren, and J.~Sun.
\newblock Deep residual learning for image recognition.
\newblock In \emph{Proceedings of the IEEE Conference on Computer Vision and Pattern Recognition (CVPR)}, 2016.

\bibitem[He et~al.(2014)He, Zhang, Ren, and Sun]{spatial_pyramid_pooling}
Kaiming He, Xiangyu Zhang, Shaoqing Ren, and Jian Sun.
\newblock Spatial pyramid pooling in deep convolutional networks for visual recognition.
\newblock \emph{CoRR}, abs/1406.4729, 2014.
\newblock URL \url{http://arxiv.org/abs/1406.4729}.

\bibitem[He et~al.(2015)He, Zhang, Ren, and Sun]{ResNet}
Kaiming He, Xiangyu Zhang, Shaoqing Ren, and Jian Sun.
\newblock Deep residual learning for image recognition.
\newblock \emph{CoRR}, abs/1512.03385, 2015.
\newblock URL \url{http://arxiv.org/abs/1512.03385}.

\bibitem[He et~al.(2017)He, Gkioxari, Doll{\'{a}}r, and Girshick]{mask_rcnn}
Kaiming He, Georgia Gkioxari, Piotr Doll{\'{a}}r, and Ross~B. Girshick.
\newblock Mask {R-CNN}.
\newblock \emph{CoRR}, abs/1703.06870, 2017.
\newblock URL \url{http://arxiv.org/abs/1703.06870}.

\bibitem[Hubert \& Arabie(1985)Hubert and Arabie]{Hubert1985_AdjustedRandIndex}
Lawrence Hubert and Phipps Arabie.
\newblock Comparing partitions.
\newblock \emph{Journal of Classification}, 2\penalty0 (1):\penalty0 193--218, 1985.

\bibitem[Jia et~al.(2023)Jia, Liu, and Huang]{jia2023_ISA_w_query_optimization}
Baoxiong Jia, Yu~Liu, and Siyuan Huang.
\newblock Improving object-centric learning with query optimization, 2023.

\bibitem[Kabra et~al.(2019)Kabra, Burgess, Matthey, Kaufman, Greff, Reynolds, and Lerchner]{multiobjectdatasets19}
Rishabh Kabra, Chris Burgess, Loic Matthey, Raphael~Lopez Kaufman, Klaus Greff, Malcolm Reynolds, and Alexander Lerchner.
\newblock Multi-object datasets.
\newblock https://github.com/deepmind/multi-object-datasets/, 2019.

\bibitem[Kim et~al.(2023)Kim, Choi, Choi, and Kim]{kim2023shepherding}
Jinwoo Kim, Janghyuk Choi, Ho-Jin Choi, and Seon~Joo Kim.
\newblock Shepherding slots to objects: Towards stable and robust object-centric learning, 2023.

\bibitem[Krizhevsky et~al.(2012)Krizhevsky, Sutskever, and Hinton]{Krizhevsky:12}
A.~Krizhevsky, I.~Sutskever, and G.~E. Hinton.
\newblock {ImageNet} classification with deep convolutional neural networks.
\newblock In F.~Pereira, C.~J.~C. Burges, L.~Bottou, and K.~Q. Weinberger (eds.), \emph{Advances in Neural Information Processing Systems 25}, pp.\  1097--1105. Curran Associates, Inc., 2012.

\bibitem[Kulkarni et~al.(2016)Kulkarni, Narasimhan, Saeedi, and J.~Tenenbaum]{Kulkarni:16}
T.~D. Kulkarni, K.~Narasimhan, A.~Saeedi, and Josh J.~Tenenbaum.
\newblock Hierarchical deep reinforcement learning: Integrating temporal abstraction and intrinsic motivation.
\newblock In D.~Lee, M.~Sugiyama, U.~Luxburg, I.~Guyon, and R.~Garnett (eds.), \emph{Advances in Neural Information Processing Systems}, volume~29, pp.\  3675--3683. Curran Associates, Inc., 2016.

\bibitem[LeCun et~al.(2004)LeCun, Huang, and Bottou]{LeCun:04}
Y.~LeCun, F.-J. Huang, and L.~Bottou.
\newblock Learning methods for generic object recognition with invariance to pose and lighting.
\newblock In \emph{Proceedings of the IEEE Conference on Computer Vision and Pattern Recognition (CVPR)}. IEEE Press, 2004.

\bibitem[Lee et~al.(1999)Lee, Itti, Koch, and Braun]{Lee:99}
Daniel~K. Lee, Laurent Itti, Christof Koch, and Jochen Braun.
\newblock Attention activates winner-take-all competition among visual filters.
\newblock \emph{Nat Neurosci}, 2\penalty0 (4):\penalty0 375--381, 1999.
\newblock \doi{10.1038/7286}.

\bibitem[Li et~al.(2006)Li, Walsh, and Littman]{Li:06}
L.~Li, T.~J. Walsh, and M.~L. Littman.
\newblock Towards a unified theory of state abstraction for {MDPs}.
\newblock In \emph{Ninth International Symposium on Artificial Intelligence and Mathematics (ISAIM)}, 2006.

\bibitem[Lin et~al.(2020)Lin, Wu, Peri, Sun, Singh, Deng, Jiang, and Ahn]{SPACE_method}
Zhixuan Lin, Yi{-}Fu Wu, Skand~Vishwanath Peri, Weihao Sun, Gautam Singh, Fei Deng, Jindong Jiang, and Sungjin Ahn.
\newblock {SPACE:} unsupervised object-oriented scene representation via spatial attention and decomposition.
\newblock \emph{CoRR}, abs/2001.02407, 2020.
\newblock URL \url{http://arxiv.org/abs/2001.02407}.

\bibitem[Lippmann(1987)]{Lippmann:87}
R.~Lippmann.
\newblock An introduction to computing with neural nets.
\newblock \emph{IEEE ASSP Magazine}, 4\penalty0 (2):\penalty0 4--22, 1987.
\newblock \doi{10.1109/MASSP.1987.1165576}.

\bibitem[Lloyd(1982)]{K_means_clustering}
S.~Lloyd.
\newblock Least squares quantization in pcm.
\newblock \emph{IEEE Transactions on Information Theory}, 28\penalty0 (2):\penalty0 129--137, 1982.
\newblock \doi{10.1109/TIT.1982.1056489}.

\bibitem[Locatello et~al.(2020)Locatello, Weissenborn, Unterthiner, Mahendran, Heigold, Uszkoreit, Dosovitskiy, and Kipf]{SlotAttention}
Francesco Locatello, Dirk Weissenborn, Thomas Unterthiner, Aravindh Mahendran, Georg Heigold, Jakob Uszkoreit, Alexey Dosovitskiy, and Thomas Kipf.
\newblock Object-centric learning with slot attention.
\newblock \emph{CoRR}, abs/2006.15055, 2020.
\newblock URL \url{https://arxiv.org/abs/2006.15055}.

\bibitem[Maass(1999)]{Maass:99}
Wolfgang Maass.
\newblock Neural computation with winner-take-all as the only nonlinear operation.
\newblock 12, 1999.
\newblock URL \url{https://proceedings.neurips.cc/paper_files/paper/1999/file/1c54985e4f95b7819ca0357c0cb9a09f-Paper.pdf}.

\bibitem[Maass(2000)]{Maass:00}
Wolfgang Maass.
\newblock On the computational power of winner-take-all.
\newblock \emph{Neural Comput.}, 12\penalty0 (11):\penalty0 2519–2535, nov 2000.
\newblock ISSN 0899-7667.
\newblock \doi{10.1162/089976600300014827}.
\newblock URL \url{https://doi.org/10.1162/089976600300014827}.

\bibitem[Mnih et~al.(2015)Mnih, Kavukcuoglu, Silver, Rusu, Veness, Bellemare, Graves, Riedmiller, Fidjeland, Ostrovski, Petersen, Beattie, Sadik, Antonoglou, King, Kumaran, Wierstra, Legg, , and Hassabis]{Mnih:15}
V.~Mnih, K.~Kavukcuoglu, D.~Silver, A.~A. Rusu, J.~Veness, M.~G. Bellemare, A.~Graves, M.~Riedmiller, A.~K. Fidjeland, G.~Ostrovski, S.~Petersen, C.~Beattie, A.~Sadik, I.~Antonoglou, H.~King, D.~Kumaran, D.~Wierstra, S.~Legg, , and D.~Hassabis.
\newblock Human-level control through deep reinforcement learning.
\newblock \emph{Nature}, 518\penalty0 (7540):\penalty0 529--533, 2015.
\newblock \doi{10.1038/nature14236}.

\bibitem[Paischer et~al.(2022)Paischer, Adler, Patil, Bitto-Nemling, Holzleitner, Lehner, Eghbal-Zadeh, and Hochreiter]{Paischer:22}
F.~Paischer, T.~Adler, V.~Patil, A.~Bitto-Nemling, M.~Holzleitner, S.~Lehner, H.~Eghbal-Zadeh, and S.~Hochreiter.
\newblock History compression via language models in reinforcement learning.
\newblock In K.~Chaudhuri, S.~Jegelka, L.~Song, C.~Szepesvari, G.~Niu, and S.~Sabato (eds.), \emph{Proceedings of the 39th International Conference on Machine Learning}, pp.\  17156--17185, 2022.

\bibitem[Patil et~al.(2022)Patil, Hofmarcher, Dinu, Dorfer, Blies, Brandstetter, Arjona-Medina, and Hochreiter]{patil:22}
Vihang Patil, Markus Hofmarcher, Marius-Constantin Dinu, Matthias Dorfer, Patrick~M Blies, Johannes Brandstetter, Jos{\'e} Arjona-Medina, and Sepp Hochreiter.
\newblock Align-{RUDDER}: Learning from few demonstrations by reward redistribution.
\newblock In Kamalika Chaudhuri, Stefanie Jegelka, Le~Song, Csaba Szepesvari, Gang Niu, and Sivan Sabato (eds.), \emph{Proceedings of the 39th International Conference on Machine Learning}, volume 162 of \emph{Proceedings of Machine Learning Research}, pp.\  17531--17572. PMLR, 17--23 Jul 2022.
\newblock URL \url{https://proceedings.mlr.press/v162/patil22a.html}.

\bibitem[Patil et~al.(2023)Patil, Hofmarcher, Rumetshofer, and Hochreiter]{Patil:23}
Vihang Patil, Markus Hofmarcher, Elisabeth Rumetshofer, and Sepp Hochreiter.
\newblock Contrastive abstraction for reinforcement learning.
\newblock In \emph{NeurIPS 2023 Workshop on Generalization in Planning}, 2023.
\newblock URL \url{https://openreview.net/forum?id=oMkUQKfsCU}.

\bibitem[Pervez et~al.(2022)Pervez, Lippe, and Gavves]{pervez2022graphcut_slots}
Adeel Pervez, Phillip Lippe, and Efstratios Gavves.
\newblock Differentiable mathematical programming for object-centric representation learning, 2022.

\bibitem[Ramesh et~al.(2021)Ramesh, Pavlov, Goh, Gray, Voss, Radford, Chen, and Sutskever]{dVAE}
Aditya Ramesh, Mikhail Pavlov, Gabriel Goh, Scott Gray, Chelsea Voss, Alec Radford, Mark Chen, and Ilya Sutskever.
\newblock Zero-shot text-to-image generation.
\newblock \emph{CoRR}, abs/2102.12092, 2021.
\newblock URL \url{https://arxiv.org/abs/2102.12092}.

\bibitem[Ramsauer et~al.(2020)Ramsauer, Sch\"{a}fl, Lehner, Seidl, Widrich, Gruber, Holzleitner, Pavlovi{\'c}, Sandve, Greiff, Kreil, Kopp, Klambauer, Brandstetter, and Hochreiter]{Ramsauer:20}
H.~Ramsauer, B.~Sch\"{a}fl, J.~Lehner, P.~Seidl, M.~Widrich, L.~Gruber, M.~Holzleitner, M.~Pavlovi{\'c}, G.~K. Sandve, V.~Greiff, D.~Kreil, M.~Kopp, G.~Klambauer, J.~Brandstetter, and S.~Hochreiter.
\newblock {Hopfield} networks is all you need.
\newblock \emph{ArXiv}, 2008.02217, 2020.

\bibitem[Rand(1971)]{RandIndex}
William~M. Rand.
\newblock Objective criteria for the evaluation of clustering methods.
\newblock \emph{Journal of the American Statistical Association}, 66\penalty0 (336):\penalty0 846--850, 1971.
\newblock ISSN 01621459.
\newblock URL \url{http://www.jstor.org/stable/2284239}.

\bibitem[Ravindran \& Barto(2003)Ravindran and Barto]{Ravindran:03b}
B.~Ravindran and A.~G. Barto.
\newblock {SMDP} homomorphisms: An algebraic approach to abstraction in semi-{Markov} decision processes.
\newblock In \emph{Proc. of the 18th Int. Joint Conf. on Artificial Intelligence (IJCAI'03)}, pp.\  1011--1016, San Francisco, CA, USA, 2003. Morgan Kaufmann Publishers Inc.

\bibitem[Schmidhuber(2015)]{Schmidhuber:15}
J.~Schmidhuber.
\newblock Deep learning in neural networks: An overview.
\newblock \emph{Neural Networks}, 61:\penalty0 85--117, 2015.
\newblock \doi{10.1016/j.neunet.2014.09.003}.

\bibitem[Seitzer et~al.(2023)Seitzer, Horn, Zadaianchuk, Zietlow, Xiao, Simon-Gabriel, He, Zhang, Schölkopf, Brox, and Locatello]{seitzer2023dinosaur}
Maximilian Seitzer, Max Horn, Andrii Zadaianchuk, Dominik Zietlow, Tianjun Xiao, Carl-Johann Simon-Gabriel, Tong He, Zheng Zhang, Bernhard Schölkopf, Thomas Brox, and Francesco Locatello.
\newblock Bridging the gap to real-world object-centric learning, 2023.

\bibitem[Silver et~al.(2016)Silver, Huang, Maddison, Guez, Sifre, van~den Driessche, Schrittwieser, Antonoglou, Panneershelvam, Lanctot, Dieleman, Grewe, Nham, Kalchbrenner, Sutskever, Lillicrap, Leach, Kavukcuoglu, Graepel, and Hassabis]{Silver:16}
D.~Silver, A.~Huang, C.~J. Maddison, A.~Guez, L.~Sifre, G.~van~den Driessche, J.~Schrittwieser, I.~Antonoglou, V.~Panneershelvam, M.~Lanctot, S.~Dieleman, D.~Grewe, J.~Nham, N.~Kalchbrenner, I.~Sutskever, T.~P. Lillicrap, M.~Leach, K.~Kavukcuoglu, T.~Graepel, and D.~Hassabis.
\newblock Mastering the game of {Go} with deep neural networks and tree search.
\newblock \emph{Nature}, 529\penalty0 (7587):\penalty0 484--489, 2016.
\newblock \doi{10.1038/nature16961}.

\bibitem[Singh et~al.(2022)Singh, Deng, and Ahn]{SLATE_singh2022}
Gautam Singh, Fei Deng, and Sungjin Ahn.
\newblock Illiterate dall-e learns to compose, 2022.

\bibitem[Spelke \& Kinzler(2007)Spelke and Kinzler]{Spelke2007-SPECK-3-dev_of_5_year_old}
Elizabeth~S. Spelke and Katherine~D. Kinzler.
\newblock Core knowledge.
\newblock \emph{Developmental Science}, 10\penalty0 (1):\penalty0 89--96, 2007.

\bibitem[Srivastava et~al.(2014{\natexlab{a}})Srivastava, Hinton, Krizhevsky, Sutskever, and Salakhutdinov]{Srivastava:14}
N.~Srivastava, G.~Hinton, A.~Krizhevsky, I.~Sutskever, and R.~Salakhutdinov.
\newblock Dropout: {A} simple way to prevent neural networks from overfitting.
\newblock \emph{J. Mach. Learn. Res.}, 15:\penalty0 1929--1958, 2014{\natexlab{a}}.

\bibitem[Srivastava et~al.(2013)Srivastava, Masci, Kazerounian, Gomez, and Schmidhuber]{Srivastava:13_compete_to_compute}
Rupesh~K Srivastava, Jonathan Masci, Sohrob Kazerounian, Faustino Gomez, and J\"{u}rgen Schmidhuber.
\newblock Compete to compute.
\newblock In C.J. Burges, L.~Bottou, M.~Welling, Z.~Ghahramani, and K.Q. Weinberger (eds.), \emph{Advances in Neural Information Processing Systems}, volume~26. Curran Associates, Inc., 2013.
\newblock URL \url{https://proceedings.neurips.cc/paper_files/paper/2013/file/8f1d43620bc6bb580df6e80b0dc05c48-Paper.pdf}.

\bibitem[Srivastava et~al.(2014{\natexlab{b}})Srivastava, Masci, Gomez, and Schmidhuber]{srivastava2014understanding_locally_comp_nets}
Rupesh~Kumar Srivastava, Jonathan Masci, Faustino Gomez, and Jürgen Schmidhuber.
\newblock Understanding locally competitive networks, 2014{\natexlab{b}}.

\bibitem[Sutton et~al.(1999)Sutton, Precup, and Singh]{Sutton:99}
R.~S. Sutton, D.~Precup, and S.~P. Singh.
\newblock Between {MDPs} and {Semi-MDPs}: {A} framework for temporal abstraction in reinforcement learning.
\newblock \emph{Artificial Intelligence}, 112\penalty0 (1-2):\penalty0 181--211, 1999.

\bibitem[Szegedy et~al.(2014)Szegedy, Liu, Jia, Sermanet, Reed, Anguelov, Erhan, Vanhoucke, and Rabinovich]{InceptionNet}
Christian Szegedy, Wei Liu, Yangqing Jia, Pierre Sermanet, Scott~E. Reed, Dragomir Anguelov, Dumitru Erhan, Vincent Vanhoucke, and Andrew Rabinovich.
\newblock Going deeper with convolutions.
\newblock \emph{CoRR}, abs/1409.4842, 2014.
\newblock URL \url{http://arxiv.org/abs/1409.4842}.

\bibitem[Vaswani et~al.(2017)Vaswani, Shazeer, Parmar, Uszkoreit, Jones, Gomez, Kaiser, and Polosukhin]{Transformer_Vaswani}
Ashish Vaswani, Noam Shazeer, Niki Parmar, Jakob Uszkoreit, Llion Jones, Aidan~N. Gomez, Lukasz Kaiser, and Illia Polosukhin.
\newblock Attention is all you need.
\newblock \emph{CoRR}, abs/1706.03762, 2017.
\newblock URL \url{http://arxiv.org/abs/1706.03762}.

\bibitem[Vezhnevets et~al.(2017)Vezhnevets, Osindero, Schaul, Heess, Jaderberg, Silver, and Kavukcuoglu]{Vezhnevets:17}
A.~S. Vezhnevets, S.~Osindero, T.~Schaul, N.~Heess, M.~Jaderberg, D.~Silver, and K.~Kavukcuoglu.
\newblock {FeUdal} networks for hierarchical reinforcement learning.
\newblock \emph{arXiv}, abs/1703.01161, 2017.

\bibitem[Wang \& Tan(2014)Wang and Tan]{wang2014_face_more_abstract_features}
Dong Wang and Xiaoyang Tan.
\newblock Unsupervised feature learning with c-svddnet, 2014.

\bibitem[Wang et~al.(2024)Wang, Zhang, Su, and Zhu]{wang2024LifeLongLearningReview}
Liyuan Wang, Xingxing Zhang, Hang Su, and Jun Zhu.
\newblock A comprehensive survey of continual learning: Theory, method and application, 2024.

\bibitem[Watters et~al.(2019)Watters, Matthey, Burgess, and Lerchner]{watters2019SBD}
Nicholas Watters, Loic Matthey, Christopher~P. Burgess, and Alexander Lerchner.
\newblock Spatial broadcast decoder: A simple architecture for learning disentangled representations in vaes, 2019.

\bibitem[Widrich et~al.(2021)Widrich, Hofmarcher, Patil, Bitto-Nemling, and Hochreiter]{Widrich:21}
M.~Widrich, M.~Hofmarcher, V.~P. Patil, A.~Bitto-Nemling, and S.~Hochreiter.
\newblock Modern {Hopfield} networks for return decomposition for delayed rewards.
\newblock In \emph{Deep RL Workshop at NeurIPS 2021}, 2021.
\newblock URL \url{https://openreview.net/forum?id=t0PQSDcqAiy}.

\bibitem[Wu et~al.(2023)Wu, Dvornik, Greff, Kipf, and Garg]{wu2023slotformer}
Ziyi Wu, Nikita Dvornik, Klaus Greff, Thomas Kipf, and Animesh Garg.
\newblock Slotformer: Unsupervised visual dynamics simulation with object-centric models, 2023.

\bibitem[Zimmermann et~al.(2023)Zimmermann, van Steenkiste, Sajjadi, Kipf, and Greff]{zimmermann2023_SlotSensitivity}
Roland~S. Zimmermann, Sjoerd van Steenkiste, Mehdi S.~M. Sajjadi, Thomas Kipf, and Klaus Greff.
\newblock Sensitivity of slot-based object-centric models to their number of slots, 2023.

\end{thebibliography}
